\documentclass[twocolumn]{autart}    % Enable this line and disable the 
\usepackage{graphicx}          % Include this line if your 
\usepackage{amssymb}
\usepackage{amsmath}
\usepackage{mathtools}
\usepackage{enumerate}
\usepackage{array}
\usepackage{multirow}
\usepackage{siunitx}
\usepackage{booktabs}
\usepackage{graphicx}
\usepackage{bm}  
\usepackage{color}   
\usepackage{soul} 
\usepackage{algorithm}
\usepackage{algpseudocode}  
\usepackage{lipsum}
\usepackage{booktabs}
\usepackage{natbib} 
\newtheorem{definition}{Definition}
\newtheorem{assumption}{Assumption}

\newtheorem{theorem}{Theorem}
\newtheorem{remark}{Remark}
\newtheorem{lemma}{Lemma}

\allowdisplaybreaks[4]

\begin{document}
	
	\begin{frontmatter}
		%\runtitle{Insert a suggested running title}  % Running title for regular 
		% papers but only if the title  
		% is over 5 words. Running title 
		% is not shown in output.
		
		\title{Federated Cubic Regularized Newton Learning with Sparsification-amplified Differential Privacy  } % Title, preferably not more 
		% than 10 words.
		
		\thanks{The work by W. Huo and L. Shi is supported by the Hong Kong RGC General Research Fund 16203723.}
		
		\author[HKUST]{Wei Huo}\ead{whuoaa@connect.ust.hk},    % Add the 
\author[KTH]{Changxin Liu}\ead{changxin@kth.se},               % e-mail address 
\author[STU]{Kemi Ding\thanksref{footnoteinfo}}\ead{dingkm@sustech.edu.cn},  % (ead) as shown
\author[KTH]{Karl Henrik Johansson}\ead{kallej@kth.se},
\author[HKUST]{Ling Shi}\ead{eesling@ust.hk}

\address[HKUST]{Department of Electronic and Computer Engineering, Hong Kong University of Science and Technology, Hong Kong}  % Please supply                                              
\address[KTH]{Division of Decision and Control Systems, School of Electrical Engineering and Computer Science, KTH Royal Institute of Technology, and also with Digital Futures, SE-10044 Stockholm, Sweden}             % full addresses
\address[STU]{School of System Design and Intelligent Manufacturing, Southern University of Science and Technology, Shenzhen, 518055, China}
\thanks[footnoteinfo]{Corresponding author.}       
		
		\begin{abstract}                          % Abstract of not more than 200 words.
		This paper explores the cubic-regularized Newton method within a federated learning framework while addressing two major concerns: privacy leakage and communication bottlenecks. 
We propose the Differentially Private Federated Cubic Regularized Newton (DP-FCRN) algorithm, which leverages second-order techniques to achieve lower iteration complexity than first-order methods. 
We incorporate noise perturbation during local computations to ensure privacy. 
Furthermore, we employ sparsification in uplink transmission, which not only reduces the communication costs but also amplifies the privacy guarantee. Specifically, this approach reduces the necessary noise intensity without compromising privacy protection. 
We analyze the convergence properties of our algorithm and establish the privacy guarantee.
Finally, we validate the effectiveness of the proposed algorithm through experiments on a benchmark dataset.
		\end{abstract}
		
		                          % keyword list or with the 
		% help of the Automatica 
		% keyword wizard
		
	\end{frontmatter}

\section{Introduction} \label{sec: introduction}
As big data grows and privacy concerns rise, conventional centralized methods for optimizing model parameters encounter substantial challenges.
Federated learning (FL) has emerged as a promising approach, allowing multiple devices to collaboratively optimize a shared model under the coordination of the central server, without sharing local data. 
FL has found applications in fields such as robotics~(\cite{yuan2024federated}) and autonomous driving~(\cite{nguyen2022deep}).
The prevailing FL algorithm is Fed-SGD, based on stochastic gradient descent (SGD)~(\cite{lowy2023private}).
%~\cite{shokri2015privacy, mcmahan2017communication}.
In this approach, each client trains a local model using SGD and uploads the gradient to the central server, which averages the gradients and updates the model. However, such first-order methods
%Although first-order methods like Fed-SGD are widely used, they 
suffer from slow convergence, which can hinder applications requiring fast processing, such as autonomous vehicles, where timely and accurate predictions are critical.
% impede the deployment of systems that require fast and efficient processing, such as autonomous vehicles where timely and accurate predictions are vital.
Newton's technique, a second-order method, offers faster convergence, but its integration into FL represents challenges. 
% However, incorporating second-order methods into FL is highly challenging. 
The key obstacle is the non-linear nature of aggregating solutions from local optimization problems for second-order approximations,  in contrast to the simpler gradient aggregation used in first-order methods. This complexity is evident in recent algorithms like GIANT~(\cite{maritan2023network}).

% While aggregating the local Hessians is possible in theory, uploading the Hessian matrices at each round incurs significant communication costs. Moreover, even without matrix transmission, communication efficiency remains a critical bottleneck in FL. For instance, if clients are some mobile devices, they usually have limited communication bandwidth.
% For example, the language model GPT-3 has billions of parameters~(\cite{brown2020language}), and thus, it is impractical to transmit them directly.
While aggregating local Hessians is theoretically feasible, uploading Hessian matrices at each round incurs significant communication costs. Even without transmitting matrices, communication efficiency remains a critical bottleneck in FL. For instance, mobile devices, which are commonly used as clients, often have limited communication bandwidth. 
%For example, GPT-3, with billions of parameters~(\cite{brown2020language}), cannot feasibly transmit all its parameters directly.
Traditional first-order optimization methods improve communication efficiency through techniques such as compression~(\cite{richtarik2021ef21}) and event trigger~(\cite{huo2024distributed}). 
Recent second-order methods, like federated Newton learning~(\cite{safaryan2021fednl}), incorporate contractive compression and partial participation to reduce communication costs. \textcolor{blue}{Building on this, \cite{zhang2024communication} further reduced communication rounds with lazy aggregation and enhanced convergence using cubic and gradient-regularized Newton methods.}
% As for faster second-order approaches, \cite{safaryan2021fednl} have recently introduced a family of federated Newton learning methods that facilitate general contractive compression operators for matrices and partial participation, thereby reducing communication costs.
% \textcolor{blue}{Based on this framework, \cite{zhang2024communication} further combined lazy aggregation to further reduced the communication rounds and utilize cubic Newton and gradient regularized Newton to improve the convergence performance.}
\cite{liu2023a} developed a distributed Newton's method with improved communication efficiency and achieved super-linear convergence. \cite{dal2024shed} presented a Newton-type algorithm to accelerate FL while addressing communication constraints.
%\textcolor{blue}{Moreover, \cite{zhang2024communication} required each worker to communicate compressed Hessians with the server only at some particular iterations, which saves both communication bits and rounds.}

In addition to slow convergence and communication costs, privacy leakage is another significant concern in FL. 
Simply storing data locally on clients does not guarantee adequate privacy.
Recent inference attacks~(\cite{zhu2019deep, liu2022membership}) show that sharing local model updates or gradients between clients and the server can result in privacy breaches. 
Sharing second-order information can also expose sensitive client data. For instance, \cite{yin2014accurate} showed that eigenvalues of the Hessian matrix can leak critical information from input images. \textcolor{blue}{In algorithms transmitting compressed Hessians~(\cite{safaryan2021fednl, zhang2024communication}), if compression does not significantly alter the eigenvalues, sensitive data may still be exposed.} Therefore, privacy preservation is vital in second-order FL.
% Sharing second-order information can also pose privacy concerns as it often encompasses the client's data. 
% For example, \cite{yin2014accurate} have shown that even the eigenvalues of the Hessian matrix can disclose some critical information from original input images.
% \textcolor{blue}{For some algorithms transmitting compressed Hessians~\cite{safaryan2021fednl, zhang2024communication}, if the compression operation does not significantly alter the eigenvalues of the Hessian matrices, sensitive information from the input data may still be exposed.}
% Thus, it is imperative to preserve privacy in FL. 
Differential privacy (DP), introduced by \cite{dwork2006differential}, has been the standard framework for privacy preservation due to its effectiveness in data analysis tasks.
\textcolor{blue}{For example, \cite{wang2023differential} explored the relationship between differential initial-value privacy and observability in linear dynamical systems. \cite{wang2024differentially} proposed a distributed shuffling mechanism based on the Paillier cryptosystem to enhance the accuracy-privacy trade-off in DP-preserving average consensus algorithms.}
In differentially private Fed-SGD, the gradient is typically augmented with Gaussian noise to achieve DP~(\cite{lowy2023private}).
%~(\cite{wei2020federated, truex2020ldp, girgis2021shuffled})
Due to the composition of DP, the required noise level is influenced by the number of iterations.
%The number of iterations $T$ influences the noise level at each iteration due to the composition of DP, with the noise level often being proportional to $\sqrt{T}$. 
Recently, \cite{ganesh2023faster} devised a second-order, differentially private optimization method that achieves $(\varepsilon, \delta)$-DP with utility loss $O(d/\varepsilon^{2})$ for $d$-dimensional model, which is optimal, i.e., the best achievable, for differentially private optimization~(\cite{bassily2014private, kairouz2021distributed}).
However, this method is restricted to centralized settings. 
Ensuring DP for second-order optimization in FL remains a challenge,
% We should address challenges such as integrating noise perturbation and reducing communication concurrently in federated Newton learning for each client, and enhancing the overall balance among privacy, accuracy, and communication efficiency.
requiring the integration of noise perturbation with communication-efficient methods and addressing trade-offs between privacy, accuracy, and communication across clients.

Motivated by the above observations, we aim to investigate federated Newton learning while jointly considering DP and communication issues in the algorithm design. 
Prior research predominantly considers DP and communication efficiency as separate entities~(\cite{li2022soteriafl, zhang2020private}). 
While some research has explored the joint trade-off among privacy, accuracy, and communication~(\cite{mohammadi2021differential, chen2021communication}), they tackled the communication and privacy in a cascaded fashion, i.e., their communication schemes do not directly impact privacy preservation. 
% Lang et al.~\cite{lang2023joint} recently proposed a method coined joint privacy enhancement and quantization, but they require the server and clients to share a common seed to decode messages.
In contrast, our study investigates the interplay between communication and privacy guarantees. 
\textcolor{blue}{Although some recent studies have employed compression in uplink transmission to improve the trade-off~(\cite{hu2023federated, chen2024privacy}), these approaches are limited to first-order learning with slow convergence. Besides, \cite{chen2024privacy} exclusively addressed central DP, which is less robust compared to privacy mechanisms at the client level.}
Specifically, we propose that each local machine uses a cubic regularized Newton method for model updates, incorporating noise perturbation during local computation.
In FL, where local model updates are typically sparse, we combine perturbation with random sparsification to enhance privacy.
% We then allow each client to upload a sparsified update to reduce communication costs.
% %In our second-order FL algorithm, we leverage the inherent characteristic of sparsification to amplify the privacy guarantee.
% Given that local model updates in FL are largely sparse, we propose integrating perturbation with random sparsification to boost privacy preservation.
Sparsification reduces the sensitivity of updates to raw data by
zeroing out some coordinates, thereby lowering privacy loss during communication.
We show that the noise intensity required for DP is proportional to the number of transmitted coordinates, meaning improved communication efficiency can reduce the noise without compromising privacy.
%reducing the number of retained coordinates eases the reduction in necessary noise intensity.
Furthermore, we illustrate that our algorithm's iteration complexity exhibits an exponential improvement compared to first-order methods, further reducing noise intensity and enhancing the trade-off between privacy and convergence.
%As the required noise intensity often depends on the number of iterations, this acceleration further reduces the noise intensity and enhances the balance between privacy and convergence trade-off.
\textcolor{blue}{Comparison of some related works with ours is shown in Table~\ref{tab: comp}.}
\begin{table*}[htbp] \color{blue}
\small 
\caption{\textcolor{blue}{Comparison of some existing works with ours}}
	 \centering
	 \begin{tabular}{|m{0.19\textwidth}<{\centering}|m{0.19\textwidth}<{\centering}|m{0.11\textwidth}<{\centering}|m{0.095\textwidth}<{\centering}|m{0.29\textwidth}<{\centering}|}
	 	\hline
	 	Work & Efficient communication & Optimization & DP & Impact of efficient communication on DP\\
        \hline
        \cite{bassily2014private, kairouz2021distributed} & $\times$ & First-order & Client-DP & $\times$ \\
        \hline
        \cite{chen2021communication, chen2024privacy} & Compressed vectors & First-order & Central DP & Increased compression did not improve privacy and reduced accuracy \\
        \hline
        \cite{safaryan2021fednl} & Compressed matrices & Second-order & $\times$ & $\times$ \\
 	 	\hline
        \cite{zhang2024communication} & Compressed matrices & Second-order & $\times$ & $\times$ \\
 	 	\hline
        Ours & Compressed vectors & Second-order & Client-DP & Increased compression enhances privacy and improves accuracy \\
        \hline
	 \end{tabular}
     \label{tab: comp}
\end{table*}

Our main contributions are summarized as follows:
\begin{itemize}
	\item[1)] %\textbf{Algorithm Design:} 
	We develop the DP-FCRN algorithm (\textbf{Algorithm~\ref{algo: one}}), which leverages second-order Newton methods for faster convergence. 
    We exploit noise perturbation in local computations to guarantee privacy preservation (\textbf{Algorithm~\ref{algo: two}}) and use sparsification to improve communication efficiency. \textcolor{blue}{Unlike previous studies that treat DP and communication burden separately~(\cite{li2022soteriafl, zhang2020private, mohammadi2021differential, chen2021communication}),
    %as separate goals and neglect the impact of efficient communication on privacy, 
    we use the inherent characteristic of sparsification to simultaneously enhance both privacy and communication efficiency.}
	\item[2)] %\textbf{Better Tradeoff:}
	We analyze the impact of sparsification on the privacy-accuracy trade-off. Specifically, we show that sparsification reduces the required noise intensity~(\textbf{Theorem~\ref{thm: dp}}), allowing for lower Gaussian noise while maintaining privacy. We also conduct a non-asymptotic analysis of utility loss and complexity (\textbf{Theorem~\ref{thm: convergence}}), demonstrating that the utility loss is optimal and that the {iteration} complexity improves over first-order methods.
	\item[3)] %\textbf{}
	We evaluate our method on the benchmark dataset. Experiment results show that our algorithm improves the model accuracy, and at the same time saves communication costs compared to Fed-SGD under the same DP guarantee.
\end{itemize}

The remainder of the paper is organized as follows. Preliminaries
and the problem formulation are provided in Section~\ref{sec: preliminaries}.
In Section~\ref{sec: algo}, a federated cubic regularized Newton learning algorithm with sparsification-amplified DP is proposed. 
Then, details on the DP analysis are shown in Section~\ref{sec: privacy} and the convergence analysis is presented in Section~\ref{sec: convergence}. 
In Section~\ref{sec: sim}, numerical simulations are presented to illustrate the obtained results. 
Finally, the conclusion and future research directions are discussed in Section~\ref{sec: conclusions}.

\emph{Notations:} 
Let $\mathbb{R}^{p}$ and $\mathbb{R}^{p \times q}$ represent the set of $p$-dimensional vectors and $p \times q$-dimensional matrices, respectively.
$I_{p} \in \mathbb{R}^{p \times p}$ represents a $p \times p$-dimensional identity matrix.
With any positive integer, we denote $[d]$ as the set of integers $\{1, 2, \dots, d \}$.
We use $[\cdot]_{j}$ to denote the $j$-th coordinate of a vector and $j$-th row of a matrix. 
Let $c$ represent a set of integers, and we denote $[X]_{c}$ as a vector containing elements $[X]_{j}$ for $j \in c$ if $X$ is a vector, and as a matrix with row vectors $[X]_{j}$ for $j \in c$ if $X$ is a matrix.
Let $\|\cdot\|$ be the $\ell_{2}$-norm vector norm.
For a convex and closed subset $\mathcal{X} \subseteq \mathbb{R}^{d}$, let {$\Pi_{\mathcal{X}}: \mathbb{R}^{d} \to \mathcal{X}$ be the Euclidean projection operator, given by $\Pi_{\mathcal{X}}(x)  = \arg\min_{y \in \mathcal{X}}\| y - x\|$}.
We use $\mathbb{P}\{\mathcal{A}\}$ to represent the probability of an event $\mathcal{A}$, and $\mathbb{E}[x]$ to be the expected value of a random variable $x$.

%\emph{Complexity Terminology:} 
The notation $O(\cdot)$ is used to describe the asymptotic upper bound. Mathematically, $h(n) = O(g(n))$ if there exist positive constants $C$ and $n_{0}$ such that $0 \leq h(n) \leq Cg(n)$ for all $n \geq n_{0}$. Similarly, the notation $\Omega(\cdot)$ provides the asymptotic lower bound, i.e., $h(n) = \Omega(g(n))$ if there exist positive constants $C$ and $n_{0}$ such that $0 \leq Cg(n) \leq h(n)$ for all $n \geq n_{0}$.
The notation $\tilde{O}(\cdot)$ is a variant of $O(\cdot)$ that ignores logarithmic factors, that is, $h(n) = \tilde{O}(g(n))$ is equivalent to $h(n) = O\left(g(n)log^{k}n\right)$ for some $k > 0$.
The notation $\Theta(\cdot)$ is defined as the tightest bound, i.e., $h(n)$ is said to be $\Theta(g(n))$ if $h(n) = O(g(n))$ and $h(n) = \Omega(g(n))$.

\section{Preliminaries and Problem Formulation} \label{sec: preliminaries}
This section introduces the fundamental setup of FL along with key concepts on Newton's methods with cubic regularization and DP. 
Subsequently, we outline the considered problem.

\subsection{Basic Setup}
We consider a federated setting with $n$ clients and a central server. 
Each client $i \in [n]$ possesses a private local dataset $\zeta_{i} = \{ \zeta_{i}^{(1)}, \dots, \zeta_{i}^{(m)} \}$ containing a finite set of $m$ data samples.
Moreover, each client has a private local cost function $f_{i}(x) = \frac{1}{m}\sum_{j=1}^{m} l(x, \zeta_{i}^{(j)})$, where $l(x, \zeta_{i}^{(j)})$ is the loss of model $x$ over the data instance $\zeta_{i}^{(j)}$ for $j \in [m]$.
With the coordination of the central server, all clients aim to train a global model $x$ by solving the following problem while maintaining their data locally:
\begin{equation} \label{eq: problem}
	\min_{x \in \mathcal{X}} \ f(x) = \frac{1}{n} \sum_{i=1}^{n} f_{i}(x),
\end{equation}
where $\mathcal{X} \subseteq \mathbb{R}^{d}$ is a convex and closed box constraint.
Specifically, the model training process takes place locally on each client, and only the updates are sent to the server for aggregation and global updates.
The optimal model parameter is defined as $x^{*} = \arg\min_{x \in \mathcal{X}} f(x)$. 

%We make the following assumptions on \textcolor{blue}{the optimization problem~\eqref{eq: problem}}:
\begin{assumption} \label{assum: model}
    The optimization problem (1) satisfies the following conditions:
    \begin{enumerate}[(\romannumeral1)]
        \item \label{assum: set} The parameter set $\mathcal{X}$ has finite diameter $D$.
        \item \label{assum: loss} The loss function $l(\cdot, \zeta)$ is $L_{0}$-Lipschitz, $L_{1}$-smooth, and has an $L_{2}$-Lipschitz Hessian for any $\zeta$ \textcolor{blue}{over $\mathcal{X}$}. 
        \item \label{assum: stro_con} The loss function $l(\cdot, \zeta)$ is $\mu$-strongly convex for any $\zeta$ \textcolor{blue}{over $\mathcal{X}$}.
    \end{enumerate}
\end{assumption}
% \begin{assumption} \label{assum: loss}
% 	$l(\cdot, \zeta)$ is $L_{0}$-Lipschitz, $L_{1}$-smooth, and has $L_{2}$-Lipschitz Hessian for any $\zeta$. 
% \end{assumption}
% \begin{assumption} \label{assum: stro_con}
% 	$l(\cdot, \zeta)$ is $\mu$-strongly convex for any $\zeta$.
% \end{assumption}
% \begin{assumption} \label{assum: set}	
% 	$\mathcal{X}$ has a finite diameter $D$. 
% \end{assumption}
%From Assumptions~\ref{assum: loss} and~\ref{assum: stro_con}, 
From Assumption~\ref{assum: model}, we infer that also $f_{i}(\cdot)$ and $f(\cdot)$ are $\mu$-strongly convex, $L_{0}$-Lipschitz, $L_{1}$-smooth, and have $L_{2}$-Lipschitz Hessian \textcolor{blue}{over $\mathcal{X}$}.

\subsection{Newton Methods with Cubic Regularization}
Newton methods~(\cite{boyd2004convex}) iteratively minimize a quadratic approximation of the function $f(\cdot)$ as 
%given by $x_{t+1} = \textcolor{blue}{\Pi_{\mathcal{X}}[x_{t} + y_{t}]}$, where
% \begin{equation} \label{eq: newton}
% \begin{aligned}
% 	y_{t} &= \arg\min_{y} \left\{ f(x_{t}) + \left< \nabla f(x_{t}), y \right> + \frac{1}{2} \left< \nabla^{2}f(x_{t})y,y \right> \right \} \\
% 	 &= -\left( \nabla^{2}f(x_{t}) \right)^{-1}\nabla f(x_{t}).
% \end{aligned}
% \end{equation}
\begin{equation} \label{eq: newton}
\begin{aligned}
 x_{t+1} = \arg\min_{x \in \mathcal{X}}  \Big\{ & f(x_{t}) + \left< \nabla f(x_{t}), x - x_{t} \right> \\
 & + \frac{1}{2} \left< \nabla^{2}f(x_{t})(x - x_{t}), x - x_{t} \right> \Big \}. %\\
	%  &= -\left( \nabla^{2}f(x_{t}) \right)^{-1}\nabla f(x_{t}).
\end{aligned}
\end{equation}
The Hessian matrix $\nabla^{2}f(x_{t})$ provides curvature information about $f(\cdot)$ at $x_{t}$. 
Newton's methods significantly improve the convergence speed of gradient descent by automatically adjusting the step size along each dimension based on the local curvature at each step.

The cubic regularized Newton method, initially introduced by~\cite{nesterov2006cubic}, incorporates a second-order Taylor expansion with a cubic regularization term. 
In particular, the update is 
% \begin{equation} \label{eq: cr}
% \begin{aligned}
% 	y_{t} = \arg\min_{y} & \bigg\{  f(x_{t}) + \left< \nabla f(x_{t}), y \right> + \frac{1}{2} \left< \nabla^{2}f(x_{t})y,y \right> \\
% 	& \quad \quad + \frac{L_{2}}{6}\| y \|^{3} \bigg \}.
% \end{aligned}
% \end{equation}
\begin{equation}
\begin{aligned} 
	x_{t+1} = \arg\min_{x \in \mathcal{X}}  \bigg\{ &  f(x_{t}) + \left< \nabla f(x_{t}), x - x_{t} \right>  \\
 &+ \frac{1}{2} \left< \nabla^{2}f(x_{t})(x - x_{t}),x - x_{t} \right>  \\
 &  + \frac{L_{2}}{6}\| x - x_{t} \|^{3} \bigg \},
\end{aligned}
\label{eq: cr}
\end{equation}
where $L_{2}$ is the Lipschitz Hessian constant in Assumption~\ref{assum: loss}.
The cubic upper bound of $f(x_{t})$ in~\eqref{eq: cr} serves as a universal upper bound regardless of the specific characteristics of the objective function. 
%It can promote global convergence }. 
However, 
%Different from~\eqref{eq: newton}, 
the function to minimize in each step of~\eqref{eq: cr} does not have a closed-form solution and it is limited to a centralized single node setting, which our algorithm addresses in a federated setting as discussed in Section~\ref{sec: algo}.
\textcolor{blue}{Cubic regularization ensures globally convergent second-order optimization with adaptive step control, avoiding the instability and exact Hessian inversion requirements of Newton’s methods while maintaining efficiency in non-convex or distributed settings.}

\subsection{Threat Model and DP}
Local datasets contain sensitive user information.
If problem~\eqref{eq: problem} is addressed in an insecure environment, information leakage could jeopardize both personal and property privacy. 
This paper considers the following adversary model:
%~(\cite{lyu2020threats}):
\begin{definition}
	(Adversary Model) 
	Adversaries can be
	\begin{itemize}
	\item[\romannumeral1)] an honest-but-curious central server that follows the protocol but may attempt to infer private client information from the received messages.
	\item[\romannumeral2)] colluding clients or clients collaborating with the central server to deduce private information about other legal clients.
	\item[\romannumeral3)] an outside eavesdropper who intercepts all transmitted messages without actively destroying communication.
	\end{itemize}
\end{definition}
Our adversary model is much stronger than some works that require a trusted third party~(\cite{hao2019efficient, chen2024privacy}).

DP is a widely used concept for quantifying privacy risk. 
It ensures that the presence or absence of any individual in a dataset cannot be inferred from the output of a randomized algorithm $\mathcal{A}$~(\cite{dwork2006differential}).
Below, we present the formal definition of DP within the context of FL.

\begin{definition} \label{defn: DP}
	 (($\varepsilon, \delta$)-DP)
% \textcolor{blue}{Consider each node has a local dataset $\zeta_{i}$.}
	The algorithm $\mathcal{A}$ is called $(\varepsilon, \delta)$-DP, if for any neighboring dataset pair {$\zeta = \cup_{i\in [n]}\zeta_{i}$ and  $\zeta^{\prime} = \cup_{i\in [n]}\zeta_{i}^{\prime}$} that differ in one data instance
 %$\zeta = \{\zeta_{1}, \dots, \zeta_{i}, \dots, \zeta_{n} \}$, $\zeta^{\prime} = \{\zeta_{1}, \dots, \zeta_{i}^{\prime}, \dots, \zeta_{n} \}$, 
 and every measurable $\mathcal{O} \subseteq \emph{Range}(\mathcal{A})$
	\footnote{$\text{Range}(\mathcal{A})$ denotes the set of all possible observation sequences under the algorithm $\mathcal{A}$.}, the output distribution satisfies
	\begin{equation} \label{eq: DP_defn}
		\mathbb{P}\{ \mathcal{A}(\zeta) \in \mathcal{O} \} \leq e^{\varepsilon} \mathbb{P}\{ \mathcal{A}(\zeta^{\prime}) \in \mathcal{O} \} + \delta,
	\end{equation}
	where the probability $\mathbb{P}\{ \cdot \}$ is taken over the randomness of $\mathcal{A}$.
\end{definition}
Definition~\ref{defn: DP} states that the output distributions of neighboring datasets exhibit small variation. 
The factor $\varepsilon$ in~\eqref{eq: DP_defn} represents the upper bound of privacy loss by algorithm $\mathcal{A}$, and $\delta$ denotes the probability of breaking this bound. Therefore, a smaller $\varepsilon$ corresponds to a stronger privacy guarantee.
\textcolor{blue}{Both Laplace and Gaussian noise can achieve DP. However, Gaussian noise, with its more concentrated distribution and superior composition properties, offers a better balance between privacy and accuracy. Therefore, we focus on the Gaussian mechanism in this work.}

\begin{lemma} \label{lem: GM} 
	(Gaussian Mechanism~(\cite{balle2018improving}))
	A Gaussian mechanism $\mathcal{G}$ for a vector-valued computation $r: \zeta \to \mathbb{R}^{d}$ is obtained by computing the function $r$ on the input data $\zeta_{i} \in \zeta$ and then adding random Gaussian noise perturbation $\nu \sim \mathcal{N}(0, \sigma^{2}I_{d})$ to the output, i.e, 
	 %Consider the Gaussian mechanism for answering the query $r: \zeta \to \mathbb{R}^{d}$:
	\begin{equation*}
		\mathcal{G} = r(\zeta) + \nu.
	\end{equation*}	
	%where $\nu \sim \mathcal{N}(0, \sigma^{2}I_{d})$.
	 %is the noise perturbation. 
	 The Gaussian mechanism $\mathcal{G}$ is $\left(\frac{\sqrt{2 \log (1.25/ \delta) }\Delta}{\sigma}, \delta \right)$-DP for any neighboring dataset $\zeta$ and $\zeta^{\prime}$, where $\Delta$ denotes the sensitivity of $r$, i.e., $\Delta = \sup_{\zeta, \zeta^{\prime}} \| r(\zeta) - r(\zeta^{\prime}) \|$. 
\end{lemma}
Lemma~\ref{lem: GM} indicates that achieving $(\varepsilon, \delta)$-DP requires adjusting the noise intensity based on the privacy guarantee $\varepsilon$ and $\delta$, as well as the sensitivity $\Delta$.

% {\color{blue} For the definition of neighboring datasets, is $\zeta_{i}^{\prime}$ also a dataset or a sample? It is used to denote the local dataset of node $i$ before.}
% \textcolor{blue}{$\zeta_{i}^{\prime}$ is a local dataset.}

\subsection{Problem Statement} \label{subsec: pro}
This paper aims to answer the following questions:
\begin{itemize}
\item[(a)] How can we develop a cubic regularized Newton algorithm for solving~\eqref{eq: problem} in a federated setting? 
\item [(b)] Can we explore the sparsification scheme to reduce communication costs while amplifying the privacy guarantee, i.e., achieving a smaller $\varepsilon$ given $\sigma$ or requiring a smaller $\sigma$ given $\varepsilon$?
\item[(c)] What level of noise intensity, i.e., $\sigma$, is necessary to attain $(\varepsilon, \delta)$-DP in the proposed algorithm?
\item[(d)] Is it possible to attain the best achievable utility loss under DP, i.e., $f(x_{T}) - f(x^{*}) = O(d/\varepsilon^{2})$ with the output $x_{T}$? 
% optimization loss? 
If achievable, what is the iteration complexity for achieving this optimal utility loss?
\end{itemize}

\section{Main Algorithm} \label{sec: algo}
In this section, we present Algorithms~\ref{algo: one} and~\ref{algo: two} to answer problems (a) and (b) in Section~\ref{subsec: pro}. 

In general, there are two approaches for integrating sparsification and privacy in FL: (1) perturb first, then sparsify, and (2) sparsify first, then perturb. 
The first approach is direct and adaptable since sparsification preserves DP and integrates smoothly with all current privacy mechanisms. 
However, in the second approach, perturbation may compromise the communication savings achieved through sparsification. Furthermore, empirical observations suggest that the first approach outperforms the second one in some scenarios~(\cite{ding2021differentially}). 
Therefore, we adopt the first approach in this study.

%\begin{figure}[t]
	%\removelatexerror
	\begin{algorithm}[t]
		\caption{%Differentially Private Federated Cubic-regularized Newton Learning (DP-FCRN)
		DP-FCRN}
		\begin{algorithmic}[1]
			\State Input: Clients' data $\zeta_{1}, \dots, \zeta_{n}$, 
			sparsification parameter $k$, DP parameters $(\varepsilon, \delta)$, and step size $\alpha$.
			\State Initialization:  Model parameter $x_{0}$.
			\For {$t = 0, 1, \dots, T-1$}
			\State $\blacktriangleright$ \textbf{Server broadcasts}
			\State Broadcast $x_{t}$ to all clients 
			\State $\blacktriangleright$ \textbf{Clients update and upload}
			\For {each client $i \in [n]$ in parallel}
			\State Sample $\zeta_{i,t}$ uniformly from $\{\zeta_{i}^{(1)}, \dots, \zeta_{i}^{(m)} \}$ and compute the local estimate gradient $\hat{g}_{i,t}= \nabla l(x_{t}, \zeta_{i,t})$ and the local estimate Hessian $\hat{H}_{i,t}= \nabla^{2} l(x_{t}, \zeta_{i,t})$ 
			\State $x_{i,t+1} = \text{GMSolver}(x_{t}, \hat{g}_{i,t}, \hat{H}_{i,t}, \tau, \sigma)$
			%\STATE Create a new random sparsifier $\mathcal{S}_{i}^{t}(\cdot)$
			\State $y_{i,t} \leftarrow \alpha (x_{i,t+1} - x_{t})$ and upload $\mathcal{S}(y_{i,t})$ to the server
			\EndFor 
			\State $\blacktriangleright$ \textbf{Server updates}
			\State $x_{t+1} = x_{t} + \frac{1}{n}\sum_{i \in \mathcal{I}_{t}} \mathcal{S}(y_{i,t})$
			\EndFor
		\end{algorithmic}
   \label{algo: one}
	\end{algorithm}
%\end{figure} 
As shown in Algorithm~\ref{algo: one}, during iteration $t$, the server broadcasts the parameter $x_{t}$ to the clients. 
Then, 
%\textcolor{blue}{to reduce the burden for computing full gradient and Hessian},
client $i$ randomly samples a data {instance} $\zeta_{i,t} \in \zeta_{i}$, estimates the local gradient $\hat{g}_{i,t} = \nabla l(x_{t}, \zeta_{i,t})$ and the local Hessian $\hat{H}_{i,t} = \nabla^{2} l(x_{t}, \zeta_{i,t})$ using its local data to minimize a local cubic-regularized upper bound of its loss function, and then does the following update
\begin{equation} \label{eq: local_comp}
% \begin{aligned}
% 	 x_{i,t+1}  =& \arg\min_{x \textcolor{blue}{ \in \mathcal{X}}}  \bigg\{  f_{i}(x_{t}) + \left< \nabla f_{i}(x_{t}), x - x_{t} \right> \\
% 	&  + \frac{1}{2} \left< \nabla^{2}f(x_{t})x - x_{t},x - x_{t} \right>   + \frac{L_{2}}{6}\| x - x_{t} \|^{3} \bigg \}.
% \end{aligned}
\begin{aligned}
	 x_{i,t+1}  =& \arg\min_{x  \in \mathcal{X}}  \bigg\{  f_{i}(x_{t}) + \left<  \hat{g}_{i,t}, x - x_{t} \right> \\
	&  + \frac{1}{2} \left< \hat{H}_{i,t}{(x - x_{t})},x - x_{t} \right>   + \frac{L_{2}}{6}\| x - x_{t} \|^{3} \bigg \}.
\end{aligned}
\end{equation}
As there is no closed form for optimal solution to~\eqref{eq: local_comp}, the client instead employs the gradient descent method to compute $x_{i,t+1}$.
To privately minimize the local cubic upper bound, Gaussian noise is added to perturb the gradient. 
This local solver utilizing the Gaussian mechanism is denoted GMSolver and is detailed in Algorithm~\ref{algo: two}.
%\begin{figure}[t]
	%\removelatexerror
	\begin{algorithm}[t]
		\caption{%Solver with Gaussian Mechanism (GMSolver)
		GMSolver}
		\begin{algorithmic}[1] 
			\State Input: Initialization $\theta_{0}$, gradient $g$, Hessian $H$, the number of iterations $\tau$, and the noise parameter $\sigma$. 
			%\ENSURE  Model parameter $x_{0}$.
			\For {$s = 0, 1, \dots, \tau - 1$}
			\State $\eta_{s} = \frac{2}{\mu(s+2)}$
			\State $\text{grad}_{s} = g + H(\theta_{s} - \theta_{0}) + \frac{L_{2}}{2}\| \theta_{s} - \theta_{0} \|(\theta_{s} - \theta_{0})$
			\State $\theta_{s+1} = \Pi_{\mathcal{X}}\left[ \theta_{s} - \eta_{s}(\text{grad}_{s} + b_{s}) \right]$, where $b_{s} \sim N(0, \sigma^{2}I_{d})$
			\EndFor 
			\State Return $\sum_{s=0}^{\tau-1} \frac{2(s+1)}{\tau(\tau+1)} \theta_{s}$ 
		\end{algorithmic}
  \label{algo: two}
	\end{algorithm}
%\end{figure} 

Following the update of the local model parameter, each client uploads its model update $x_{i,t+1} - x_{t}$ to the server.
To address the communication challenges in uplink transmissions, the random-$k$ sparsifier is employed to reduce the size of the transmitted message by a factor of $k/d$~(\cite{li2021canita}):
\begin{definition}
	(Random-$k$ Sparsification): For $x \in \mathbb{R}^{d}$ and a parameter $k \in [d]$, the random-$k$ sparsification operator is
	\begin{equation*}
		\mathcal{S}(x) := \frac{d}{k}(\xi_{k} \odot x),
	\end{equation*}
	where $\xi_{k} \in \{0,1 \}^{d}$ is a uniformly random binary vector with $k$ nonzero entries, i.e., $\| \xi_{k} \|_{0} = k$ and $\odot$ represents the element-wise Hadamard product. 
\end{definition}

Integrating private GMSolver and random-$k$ sparsification, the proposed algorithm simultaneously addresses privacy preservation and communication efficiency as depicted in Algorithm~\ref{algo: one}. 
A scaling factor $\alpha>0$ is introduced for convergence analysis.

\begin{remark}
	As pointed out by~\cite{lacoste2012simpler}, the output of Algorithm~\ref{algo: two} can be computed online. 
	Specifically, setting $z_{0} = \theta_{0}$, and recursively defining $z_{s} = \rho_{s}\theta_{s} + (1-\rho_{s})z_{s-1}$ for $s \geq 1$, with $\rho_{s} = \frac{2}{s+1}$. It is a straightforward calculation to check that $z_{\tau} = \sum_{s=0}^{\tau-1} \frac{2s}{\tau(\tau+1)} \theta_{s}$.
\end{remark}

\begin{remark}
Solving \eqref{eq: cr} directly requires significant computational resources. To improve efficiency, we use parallel cooperative solving across multiple clients. Since \eqref{eq: local_comp} lacks a closed-form solution, we propose a local training approach for its resolution. Privacy is preserved through noise perturbation during local training, while sparsification in uplink transmission reduces communication costs and further enhances privacy protection. \textcolor{blue}{Unlike existing methods that transmit compressed Hessian matrices \cite{safaryan2021fednl, zhang2024communication}, we transmit compressed vectors to further minimize communication overhead.}
   % Directly solving \eqref{eq: cr} requires substantial computational resources. Therefore, we leverage parallel cooperative solving by multiple clients to enhance learning efficiency. To address the absence of a closed-form solution in \eqref{eq: local_comp}, we propose a local training approach for its resolution. Privacy preservation is ensured through noise perturbation during local training, while sparsification in uplink transmission not only reduces communication costs but also enhances privacy protection.
   % \textcolor{blue}{Different from existing works that transmitted compressed Hessian matrices~(\cite{safaryan2021fednl, zhang2024communication}), we transmitted compressed vectors to further avoid high communication overload.}
\end{remark}

\section{Privacy Analysis} \label{sec: privacy}
In this section, we prove the privacy guarantee provided by Algorithm~\ref{algo: one}. 
To facilitate privacy analysis, we make the following assumption.
\begin{assumption} \label{assum: bounded} 
    For any data sample $\zeta_{i}^{(j)} \in \zeta_{i}$ and $h \in [d]$, we have
    \begin{equation*}
      \left  | \left[ \nabla l(x, \zeta_{i}^{(j)}) \right]_{h} \right | \leq \frac{\textcolor{blue}{L_{0}}}{\sqrt{d}}, \quad
      \left \| \left[ \nabla^{2} l(x, \zeta_{i}^{(j)}) \right]_{h}  \right\| \leq \frac{\textcolor{blue}{L_{1}}}{\sqrt{d}}
      %+ \nabla^{2} l (x, \zeta_{i}^{(j)})v \right  |_{h} \leq \frac{G}{\sqrt{d}},
    \end{equation*}
    for any $x, v \in \mathcal{X}$ and $i \in [n]$.
\end{assumption}
% \begin{assum} \label{assum: bounded} \color{blue}
%     For any data sample $\zeta_{i}^{(j)} \in \zeta_{i}$ and $h \in [d]$, we have
%     \begin{equation*}
%       \left  |\nabla l(x, \zeta_{i}^{(j)}) + \nabla^{2} l (x, \zeta_{i}^{(j)})v \right  |_{h} \leq \frac{G}{\sqrt{d}},
%     \end{equation*}
%     for any $x, v \in \mathcal{X}$ and $i \in [n]$.
% \end{assum}
%From Assumption~\ref{assum: loss} and~\ref{assum: set}, there 

Assumption~\ref{assum: bounded} characterizes the sensitivity of each coordinate of the gradient $\nabla l(x, \zeta_{i}^{(j)})$ and each row of the Hessian $\nabla^{2} l(x, \zeta_{i}^{(j)})$. 
\textcolor{blue}{This assumption is crucial for analyzing the interaction between element selection via sparsification and privacy amplification, as discussed in~\cite{hu2023federated, chen2024privacy}.}
It implies that $\left\| \nabla l(x, \zeta_{i}^{(j)}) \right\| \leq \textcolor{blue}{L_{0}}$ and $\left\| \nabla^{2} l(x, \zeta_{i}^{(j)})\right\|_{2} \leq \left\| \nabla^{2} l(x, \zeta_{i}^{(j)})\right\|_{F} \leq \textcolor{blue}{L_{1}}$, \textcolor{blue}{which holds under the assumption of a bounded parameter set}.
%which can be enforced by the gradient and Hessian clip techniques~(\cite{ganesh2023faster}).

To analyze the interplay between the sparsification and privacy, let $c_{i}^{t}$ denote the randomly selected coordinate set for client $i$ at round $t$, i.e., $\mathcal{S}(\cdot) = \frac{d}{k} [\cdot]_{c_{i}^{t}}$. 

% \begin{lemma}
%     If $\text{grad}_{i}^{t,s}$ and $b_{i}^{t,s}$ are the gradient and noise used by client $i$ at iteration $s$ in GMSolver and the communication round $t$ in Algorithm~\ref{algo: one}, respectively, it is equivalent that 
% \end{lemma}

An important observation is that only the values in $c_{i}^{t}$ are transmitted to the central server, i.e., 
\begin{align*}
	\textcolor{blue}{\mathcal{S}(y_{i,t}) = \frac{d}{k} \left[ \alpha(x_{i,t+1} - x_{i,t}) \right]_{c_{i}^{t}} = \frac{\alpha d}{ k} \left( [x_{i,t+1}]_{c_{i}^{t}} - [x_{i,t}]_{c_{i}^{t}} \right).}
\end{align*}
The gradient update information is contained in $[x_{i,t+1}]_{c_{i}^{t}}$ and
\begin{equation*} 
\begin{aligned}
    [x_{i,t+1}]_{c_{i}^{t}} =& \left[ \sum_{s=0}^{\tau-1} \frac{2(s+1)}{\tau(\tau+1)} \theta_{i}^{t,s} \right]_{c_{i}^{t}} = \sum_{s=0}^{\tau-1} \frac{2(s+1)}{\tau(\tau+1)} [\theta_{i}^{t,s}]_{c_{i}^{t}},
\end{aligned}
\end{equation*}
where $\theta_{i}^{t,s}$ denotes the optimization variable used by client $i$ at iteration $s$ in Algorithm~\ref{algo: two} and the communication round $t$ in Algorithm~\ref{algo: one}.
Based on step 4 in the GMSolver, we have
\begin{align*} 
	[\theta_{i}^{t,s+1}]_{c_{i}^{t}} = \big[\Pi_{\mathcal{X}} \left[ \theta_{i}^{t,s} - \eta_{s}(\text{grad}_{i}^{t,s} + b_{i}^{t,s}) \right] \big]_{c_{i}^{t}},
\end{align*}
where $\text{grad}_{i}^{t,s}$ and $b_{i}^{t,s}$ are the gradient and noise used by client $i$ at iteration $s$ in GMSolver and the communication round $t$ in Algorithm~\ref{algo: one}, respectively.
Since projection into a box constraint does not influence the set of selected coordinators $c_{i}^{t}$, what matters in local computation is
\begin{equation*}
    \left[ \theta_{i}^{t,s} - \eta_{s}(\text{grad}_{i}^{t,s} + b_{i}^{t,s}) \right]_{c_{i}^{t}} = [\theta_{i}^{t,s}]_{c_{i}^{t}} - \eta_{s}[\text{grad}_{i}^{t,s} + b_{i}^{t,s}]_{c_{i}^{t}}.
\end{equation*}
%Therefore, it is equivalent to use 
% {\color{blue} please double check this. Why the projection can commute with coordinate selection is not straightforward to me}
% \begin{align*}
% 	[\theta_{s+1}]_{c_{i}^{t}} = \Pi_{\mathcal{X}}\left[ [\theta_{s}]_{c_{i}^{t}} - \eta_{s}[\text{grad}_{s} + b_{s}]_{c_{i}^{t}} \right]
% \end{align*}
%at step 4 in Algorithm~\ref{algo: two}.

According to the above analysis, we conclude that the crucial aspect of privacy protection lies in the sparsified noisy gradient update, which can be expressed as
\begin{equation*}
%\mathcal{S}\left( \text{grad}_{i}^{t,s} + b_{i}^{t,s} \right) 
	%=  \left( [\text{grad}_{i}^{t,s}]_{c_{i}^{t}} + [b_{i}^{t,s}]_{c_{i}^{t}} \right),
	[\text{grad}_{i}^{t,s} + b_{i}^{t,s}]_{c_{i}^{t}} = [\text{grad}_{i}^{t,s}]_{c_{i}^{t}} + [b_{i}^{t,s}]_{c_{i}^{t}}.
\end{equation*}
%for client $i$ at the round $t$, where $\text{grad}_{i}^{t,s}$ and $b_{i}^{t,s}$ are the gradient and noise used by client $i$ at iteration $s$ in Algorithm~\ref{algo: two} and the communication round $t$ in Algorithm~\ref{algo: one}, respectively.
We observe that the sparsification makes Gaussian noises only perturb the values at coordinates within $c_{i}^{t}$. 
If noise is added only at the selected coordinates, the level of privacy remains the same.
In other words, we ensure the same privacy level even when incorporating a diminished amount of additional noise, thereby enhancing the optimization accuracy.
Subsequently, we only need to analyze the privacy budget of $[\text{grad}_{i}^{t,s}]_{c_{i}^{t}}$ after adding noise $[b_{i}^{t,s}]_{c_{i}^{t}}$.

For client $i$, considering any two neighboring dataset $\zeta_{i}$ and $\zeta_{i}^{\prime}$ of the same size $m$ but with only one data sample different (e.g., $\zeta_{i}^{j_{0}}$ and $\zeta_{i}^{j_{0} \prime}$).
Denote $\Delta$ as the $\ell_{2}$-sensitivity of $[\text{grad}_{i}^{t,s}]_{c_{i}^{t}}$, and we have
\begin{align} \label{eq: sen}
	&\Delta^{2} \nonumber \\ 
	=& {\max_{\zeta, \zeta^{\prime}}} \Big \| \left[\hat{g}_{i,t}\right]_{c_{i}^{t}} - \left[\hat{g}_{i,t}^{\prime}\right]_{c_{i}^{t}} + \left[\hat{H}_{i,t}(\theta_{i}^{t,s} - x_{t}) \right]_{c_{i}^{t}} \nonumber \\
	& \quad \quad \quad- \left[\hat{H}_{i,t}^{\prime}( \theta_{i}^{t,s} - x_{t} )\right]_{c_{i}^{t}} \Big \|^{2} \nonumber \\
	=& {\max_{\zeta, \zeta^{\prime}}} \bigg\|  \left[\nabla l(x_{t}, \zeta_{i}^{j_{0}}) - \nabla l(x_{t}, \zeta_{i}^{j_{0} \prime}) \right]_{c_{i}^{t}} \nonumber \\
	& \quad \quad + \left[ (\nabla^{2} l(x_{t}, \zeta_{i}^{j_{0}}) - \nabla^{2} l(x_{t}, \zeta_{i}^{j_{0} \prime})) ( \theta_{i}^{t,s} - x_{t} ) \right]_{c_{i}^{t}} \bigg \|^{2} \nonumber \\
	%\leq & \frac{2k (L_{0} + L_{1}D)^{2} }{d},
 \leq & \frac{4k {(\textcolor{blue}{L_{0}} + \textcolor{blue}{L_{1}}D)}^{2} }{d},
\end{align}
where the last inequality holds from
\begin{equation*} 
    \begin{aligned}
    & \bigg \| \left[\nabla l(x_{t}, \zeta_{i}^{j_{0}}) - \nabla l(x_{t}, \zeta_{i}^{j_{0} \prime}) \right]_{c_{i}^{t}} \\
	& \quad \quad + \left[ (\nabla^{2} l(x_{t}, \zeta_{i}^{j_{0}}) - \nabla^{2} l(x_{t}, \zeta_{i}^{j_{0} \prime})) ( \theta_{i}^{t,s} - x_{t} ) \right]_{c_{i}^{t}} \bigg \| \\
 \leq & \left \| \left[\nabla l(x_{t}, \zeta_{i}^{j_{0}}) - \nabla l(x_{t}, \zeta_{i}^{j_{0} \prime}) \right]_{c_{i}^{t}} \right \| \\
 &+  \left\| \left[ \nabla^{2} l(x_{t}, \zeta_{i}^{j_{0}}) - \nabla^{2} l(x_{t}, \zeta_{i}^{j_{0} \prime}) \right]_{c_{i}^{t}} \left[ \theta_{i}^{t,s} - x_{t}  \right]_{c_{i}^{t}} \right \| \\
 \leq & \frac{2\sqrt{k}\textcolor{blue}{L_{0}}}{\sqrt{d}} + \frac{2\sqrt{k}\textcolor{blue}{L_{1}}D}{\sqrt{d}}.
\end{aligned}
\end{equation*}

Lemma~\ref{lem: GM} indicates that the noise intensity required to achieve $(\varepsilon, \delta)$-DP relies on the sensitivity. 
From~\eqref{eq: sen}, sparsification reduces the conventional sensitivity $2(\textcolor{blue}{L_{0}} + \textcolor{blue}{L_{1}}D)$ by a factor of $\sqrt{k/d}$, thereby decreasing sensitivity and reducing the required noise intensity.
 \textcolor{blue}{For each client's sensitive local dataset $\zeta_{i}$, $\forall i \in [n]$, if we treat DP-FCRN as the algorithm $\mathcal{A}$ defined in Definition~\ref{defn: DP}, the worst-case observation by the attacker $\mathcal{A}(\zeta_{i}) = \{\mathcal{S}(y_{i,t}) | 0 \leq t \leq T \}$.}
Theorem~\ref{thm: dp} states a sufficient condition for achieving $(\varepsilon, \delta)$-DP based on the reduced sensitivity resulting from sparsification.

\begin{theorem} \label{thm: dp}
	Suppose Assumption~\ref{assum: model} holds, and the random-$k$ sparsifier with $k \leq d$ is used in Algorithm~\ref{algo: one}.
    Given $m$, $\tau$, $\varepsilon \in (0,1]$, and $\delta_{0} \in (0,1]$, if the noise variance
	\begin{equation} \label{eq: noise}
		%\sigma^{2} \geq \frac{80 \tau T k\log(1.25/\delta_{0})(L_{0} + L_{1}D)^{2}}{ \varepsilon^{2} m^{2} d}
  \sigma^{2} \geq \frac{160 \tau T k\log(1.25/\delta_{0}){(\textcolor{blue}{L_{0}} + \textcolor{blue}{L_{1}}D)}^{2}}{ \varepsilon^{2} m^{2} d}
	\end{equation}
	and $T \geq \frac{ \varepsilon^{2}}{4 \tau}$, then DP-FCRN is $(\varepsilon, \delta)$-DP for $\zeta_{i}$, $\forall i \in [n]$, with some constant $\delta \in (0,1]$. Specifically, for any output set of DP-FCRN, $\mathcal{A}(\zeta_{i})$, we have
    \begin{equation}
        \textcolor{blue}{\mathbb{P}\{ \mathcal{A}(\zeta_{i}) \in \mathcal{O} \} \leq e^{\varepsilon} \mathbb{P}\{ \mathcal{A}(\zeta_{i}^{\prime}) \in \mathcal{O} \} + \delta.}
    \end{equation}
\end{theorem}
\begin{pf}
	The proof is provided in Appendix~\ref{app: thm_dp}.
    %The proof is provided in Appendix~B~(\cite{huo2024federated}).
\end{pf}

%\begin{rem}
%	Liu et al.~\cite{liu2024distributed} amplify the DP by randomly selecting a set of agents at each communication iteration. 
%	Although the required noise can be reduced by decreasing the number of active agents, their required iterations for DP will be increased, which would eliminate the advantage of subsampling. 
%	In our proposed algorithm, the required number of communication rounds is independent of the number of selected coordinates. 
%	Therefore, to achieve the same level of DP, the required noise under our algorithm can be reduced by decreasing $k$. In other words, the fewer transmitted bits, the less noise required for $(\varepsilon, \delta)$-DP.
%\end{rem}
\begin{remark}
	The required noise intensity is proportional to the sparsification ratio, $k/d$. Therefore, to achieve the same level of DP, the required noise under our algorithm can be reduced by decreasing $k$. In other words, the fewer transmitted bits, the less noise required for $(\varepsilon, \delta)$-DP.
    \textcolor{blue}{The assumption that $\varepsilon \in (0,1]$ is motivated by the need to ensure the validity of theoretical results, such as composition theorems and privacy amplification, which often require $\varepsilon$ to be small. Additionally, small $\varepsilon$ aligns with the goal of providing strong privacy guarantees, making this range both theoretically and practically relevant for differential privacy research.}
\end{remark}

{\color{blue}{
\begin{remark}
    Common compression methods include quantizers and sparsifiers. However, quantization can increase the sensitivity of gradient updates and disrupt the distribution of Gaussian perturbations, making both algorithm design and analysis more difficult. In contrast, sparsifiers simply set some elements to zero, reducing the sensitivity of the messages and making privacy amplification more tractable. Moreover, compared to the Top-$k$ sparsifier, the random-$k$ sparsifier introduces additional randomness, further enhancing the privacy guarantee. In the future, it will be interesting to study the potential privacy amplification under other compression schemes.
\end{remark}
}
}

\section{Convergence Analysis} \label{sec: convergence}
This section presents the convergence analysis of Algorithm~\ref{algo: one}. 
In each step of the algorithm, a global cubic upper bound function $\phi: \mathcal{X} \times \mathcal{X} \to \mathbb{R}$ for $f(w)$ is constructed as
\begin{align*}
		&\phi(v;w) \\
\triangleq & f(w) + \left< \nabla f(w), v-w \right> + \frac{1}{2}\left< \nabla^{2}f(w)(v-w), v-w \right> \\
& + \frac{M}{6}\|v-w \|^{3}, \quad \forall v \in \mathcal{X}, 
\end{align*}
and local cubic upper bound functions $\phi_{i}: \mathcal{X} \times \mathcal{X} \to \mathbb{R}$ for $f_{i}(w)$, $i \in \{1, 2, \dots, n\}$, as
\begin{align} \label{eq: phi_1}
	&\phi_{i}(v;w) \nonumber \\
\triangleq & f_{i}(w) + \left< \nabla f_{i}(w), v-w \right> + \frac{1}{2}\left< \nabla^{2}f_{i}(w)(v-w), v-w \right> \nonumber \\
& + \frac{M}{6}\|v-w \|^{3}, \quad \forall v \in \mathcal{X}. 
\end{align}

Algorithm~\ref{algo: two} uses the standard SGD to solve~\eqref{eq: local_comp} and the local cubic upper bound $\phi_{i}(x;x_{t})$ is a strongly convex function. \textcolor{blue}{Since each sample $\zeta_{i}^{(j)}$ is chosen with equal probability, i.e.,  $\mathbb{P}(\zeta_{i}^{(j)}) = 1/m$, $\mathbb{E}[ \hat{g}_{i,t} ] = \sum_{j=1}^{m} \nabla l(x, \zeta_{i}^{(j)}) \mathbb{P}(\zeta_{i}^{(j)}) = \nabla f_{i}(x)$ and $\mathbb{E}[ \hat{H}_{i,t} ] = \sum_{j=1}^{m} \nabla^{2} l(x, \zeta_{i}^{(j)}) \mathbb{P}(\zeta_{i}^{(j)})  = \nabla^{2} f_{i}(x)$ are unbiased estimates}. Therefore, we can obtain the suboptimality gap based on the typical SGD analysis.

\begin{lemma} \label{lem: solver}
	Suppose that Assumptions~\ref{assum: loss}--\ref{assum: set} hold. 
	%For every $\beta \in (0,1)$, 
 Given parameters $ \varepsilon \in  (0,1]$, $\delta_{0} \in (0,1]$, and $w \in \mathcal{X}$ the output of Algorithm~\ref{algo: two}, if we set the number of local iterations as
	\begin{equation} \label{eq: tau}
		\tau = \frac{(L_{0} + L_{1}D + MD^{2}/2)^{2}\varepsilon^{2}m^{2}}{kT \log(1/\delta_{0})(\textcolor{blue}{L_{0}} + \textcolor{blue}{L_{1}}D)^{2}},
	\end{equation}
	and the noise as~\eqref{eq: noise},
then $\hat{v}$ satisfies
	\begin{equation} \label{eq: solver_gap}
	\begin{aligned}
		&\mathbb{E}[\phi_{i,t}(\hat{v};w)] - \min_{v \in \mathcal{X}} \phi_{i,t}(v;w) \\
		=& O\left( \frac{k \log(1 / \delta_{0}) {(\textcolor{blue}{L_{0}} + \textcolor{blue}{L_{1}}D)}^{2}T}
		{\varepsilon^{2}m^{2}\mu } \right).
	\end{aligned}
	\end{equation}
\end{lemma}
\begin{pf}
	The proof is provided in Appendix~\ref{app: lem_solver}.
    %The proof is provided in Appendix~C~(\cite{huo2024federated}).
\end{pf}

Lemma~\ref{lem: solver} quantifies the suboptimal gap when solving~\eqref{eq: local_comp} with Algorithm~\ref{algo: two} for each client in every communication round. Based on this result, we are in a position to provide the convergence of DP-FCRN.
\begin{theorem} \label{thm: convergence}
Suppose that Assumptions~\ref{assum: loss}--\ref{assum: set} hold and the random-$k$ sparsifier with $k \leq d$ is used in Algorithm~\ref{algo: one}.
%Then for every $\beta \in (0,1)$, 
Given parameters $m$ and $\varepsilon \in  (0,1]$, $\delta_{0} \in (0,1]$, by setting the number of local iterations as~\eqref{eq: tau}, the step size as $\alpha >1$ and 
	\begin{equation*}
		\alpha = O \left(  \frac{k \log(1/\delta_{0}){(\textcolor{blue}{L_{0}} + \textcolor{blue}{L_{1}}D)}^{2}T}{\varepsilon^{2}m^{2}\mu (L_{0} + L_{1}D + MD^{2}/2)D} \right),
	\end{equation*}
	and the number of iterations in DP-FCRN to
	\begin{equation*}
	\begin{aligned}
		T = \Theta \Bigg( & \frac{\sqrt{L_{2}}( f(x_{0}) - f(x^{*}) )^{\frac{1}{4}}}{\mu^{\frac{3}{4}}} \\
		 & \quad \quad +  \log \log\left( \frac{\varepsilon m }{\sqrt{k\log(1/\delta_{0})}} \right)\Bigg),
	\end{aligned}
	\end{equation*}
	then the output of DP-FCRN, that is, $x_{T}$, preserves $(\varepsilon, \delta)$-DP and
	\begin{equation*}
	\begin{aligned}
		& \mathbb{E}[f(x_{T})] - f(x^{*}) \\
		\leq & \tilde{O}\left( \frac{k \log(1 / \delta_{0}) {(\textcolor{blue}{L_{0}} + \textcolor{blue}{L_{1}}D)}^{2}  }
		{\varepsilon^{2}m^{2}\mu } \cdot  \frac{\sqrt{L_{2}}( f(x_{0}) - f(x^{*}) )^{\frac{1}{4}}}{\mu^{\frac{3}{4}}}  \right).
	\end{aligned}
	\end{equation*}
\end{theorem}

\begin{pf}
	The proof is provided in Appendix~\ref{app: thm_con}.
    %The proof is provided in Appendix~D~(\cite{huo2024federated}).
\end{pf}

\begin{remark}
\textcolor{blue}{With the boundedness established in Assumption 2, existing DP algorithms for strongly convex functions achieve the best bound for optimization error, $O\left( \frac{d}{\varepsilon^{2} } \right)$~(\cite{bassily2014private, kairouz2021distributed}).} 
This indicates that the error bound derived in Theorem~\ref{thm: convergence} is optimal w.r.t. the privacy loss $\varepsilon$.
Furthermore, our result $O\left( \frac{k}{\varepsilon^{2}} \right)$ reduces the error bound by a factor $k/d$, attributed to sparsification. 
This result underscores how efficient communication better balances the trade-off between privacy and utility.
\textcolor{blue}{Unlike the recent algorithm in~\cite{chen2024privacy}, which assumes a trusted central server, we adopt a client-level differential privacy (DP) approach that offers stronger and more robust privacy protection. Furthermore, the error bound in~\cite{chen2024privacy} increases when fewer coordinates are transmitted, implying that higher communication efficiency leads to worse convergence accuracy. In contrast, the error bound of our algorithm shows that more efficient communication reduces convergence error. Thus, in the context of federated second-order learning, we are the first to improve the trade-off between privacy and accuracy through efficient communication.}
\end{remark}

\begin{remark}
While DP-FCRN does not explicitly include a switching step, the proof of Theorem~\ref{thm: convergence} indicates that DP-FCRN operates in two distinct phases. 
Initially, when $x_{t}$ is distant from $x^{*}$, the convergence rate is $1/T^{4}$. Subsequently, as $x_{t}$ approaches $x^{*}$, the algorithm transitions to the second phase with a convergence rate of $\exp(\exp(-T))$.
In summary, leveraging second-order techniques in our algorithm significantly improves the oracle complexity compared to first-order methods~(\cite{liu2024distributed}).
\end{remark}

\begin{remark}
 \textcolor{blue}{The privacy analysis is independent of convexity assumptions. While many non-convex algorithms focus on first-order stationary points, which may be poor local minima or saddle points, future work will explore convergence to second-order stationary points using cubic regularization. This can reduce the risk of saddle points and improve local minima. Additionally, time-varying step sizes will be essential for optimizing the achievable bounds.}
\end{remark}

% \begin{remark}
% \textcolor{blue}{
%     What happens if we directly use Newton?}
% \end{remark}

\section{Numerical Evaluation} \label{sec: sim}
In this section, we evaluate the effectiveness of DP-FCRN with different \textcolor{blue}{sparsification ratios} and compare them to the first-order Fed-SGD with DP~(\cite{lowy2023private}).
\subsection{Experimental Setup}
We test our algorithm on the benchmark datasets \emph{epsilon}~(\cite{sonnenburg2006large}), which include 400,000 samples and 2,000 features for each sample.
The data samples are evenly and randomly allocated among the $n=40$ clients. The clients cooperatively solve the following logistic regression problem:
\begin{equation*}
	\min_{x {\in \mathcal{X}}} f(x) = \frac{1}{n} {\sum_{i=1}^{n}} f_{i}(x), 
\end{equation*}
where
%\begin{equation*}
%	f_{i}(x) = \frac{1}{m} \sum_{j=1}^{m} h\Big(1-y_{i}^{(j)}\left<C_{i}^{(j)}, x\right>\Big) + \frac{|\zeta_{i}|}{n} \|x\|^{2} 
%\end{equation*}
\begin{equation*}
	f_{i}(x) = \frac{1}{m} \sum_{j=1}^{m} \log(1+\exp(-b_{j}a_{j}^{\top}x)) + \frac{1}{2m}\|x \|^{2},
\end{equation*}
%where $h(v) = \ln(1+e^{v})$ is the softplus function, an approximation to $\max(0, v)$. 
$\mathcal{X} \subseteq [-0.5, 0.5]^{d}$, $m$ is the number of samples in the local dataset, and $a_{j} \in \mathbb{R}^{d}$ and $b_{j} \in \{ -1, 1\}$ are the data samples.
%Note that the objective function above is slightly modified from the conventional formulation of the SVM, with the inclusion of the regularization for $x$.
%This is to ensure the validity of assumptions that each local objective function is strongly convex and has Lipshcitz continuous gradient and Hessian.

As DP parameters, we consider $\varepsilon \in \{ 0.4, 0.6, 0.8, 1\}$ and $\delta_{0}=0.01$. The random noise is generated according to~\eqref{eq: noise}.
\textcolor{blue}{We choose $\alpha = 1$. As for Fed-SGD, we set the learning rate as one. Moreover, $L_{0}=0.1$, $L_{1}=1$, $M=1$, $D=0.1$, $\delta_{0}=0.01$, and we calculate the value of $\tau$ using~\eqref{eq: tau}.}
In iteration $t$, client $i$ processes one data point from $\zeta_{i}$ and the server updates $x_{t}$ accordingly. Upon finishing processing the entire dataset, one epoch is completed. We conduct the algorithm for four epochs and repeat each experiment five times.
We show the mean curve along with the region representing one standard deviation.
The convergence performance of the algorithm is evaluated by training suboptimality and testing accuracy over iterations.
Training suboptimality is calculated by $f(x_{t}) - f(x^{*})$, where $f(x^{*})$ is obtained using the LogisticSGD optimizer from scikit-learn~(\cite{pedregosa2011scikit}).
Testing accuracy is determined by applying the logistic function to the entire dataset. It is calculated as the percentage of correct predictions out of the total number of predictions.

\subsection{Performance and Comparison with Fed-SGD}
\begin{figure}[t]  
	\centering
	\includegraphics[width=1\linewidth]{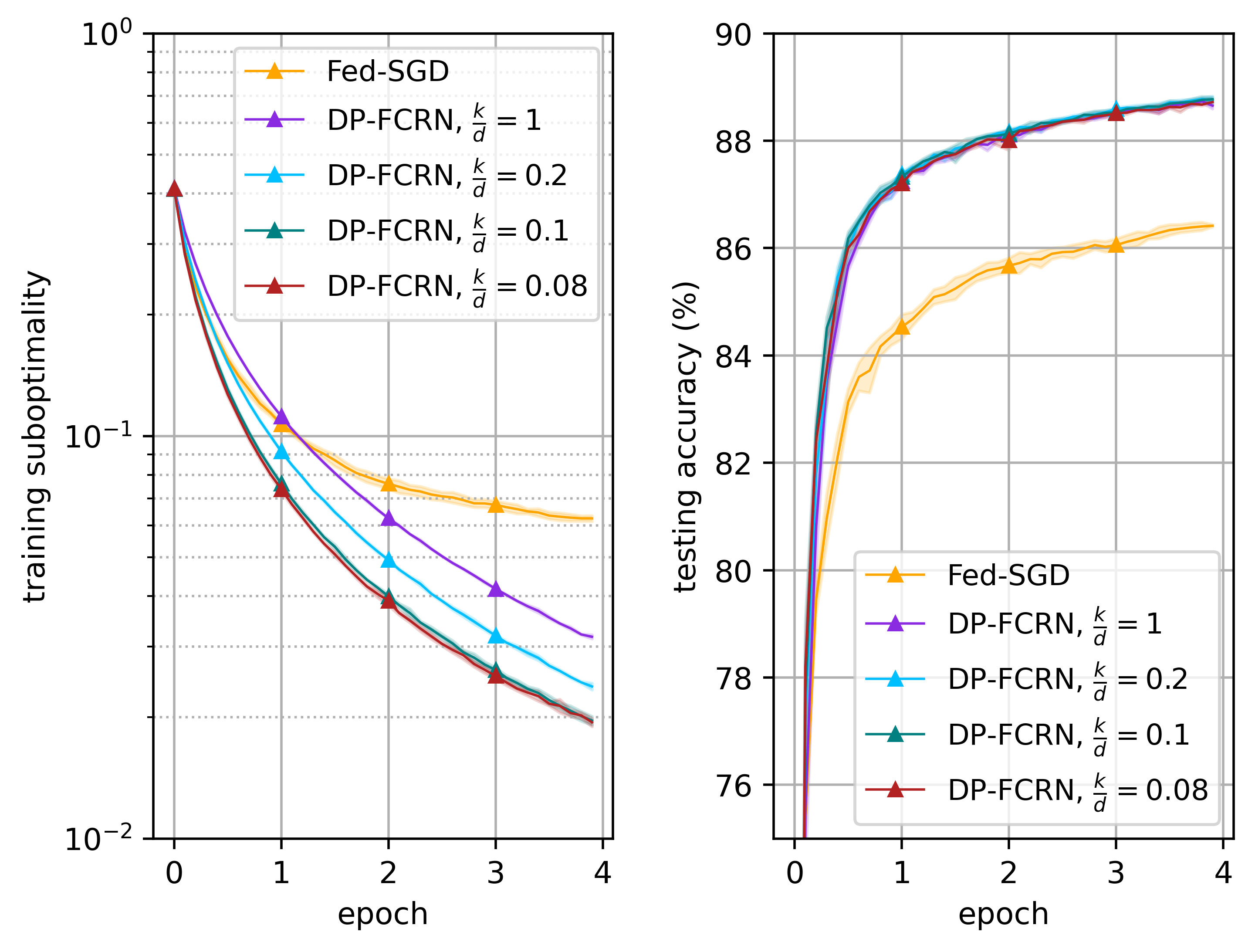}
	\caption{Performance comparison between Fed-SGD with DP and DP-FCRN with $\varepsilon = 0.8$.} 
	\label{fig: compare}
\end{figure}

By setting the privacy budget as $\varepsilon = 0.8$, we compare the convergence performance between first-order Fed-SGD with DP and Algorithm~\ref{algo: one} with different choices of sparsification ratio $k/d \in \{0.08, 0.1, 0.2, 1\}$.
Fig.~\ref{fig: compare} implies that DP-FCRN outperforms Fed-SGD with DP in terms of optimization accuracy and convergence speed.
Moreover, employing a larger sparsification ratio $k/d$ in DP-FCRN results in worse training suboptimality, verifying Theorem~\ref{thm: convergence}. 
We find that keeping more coordinates in sparsification leads to more complete information transmission together with increased noise. 
The results shown in Fig.~\ref{fig: compare} indicate that, in certain settings, the benefit of noise reduction for convergence performance may outweigh the negative impacts arising from information completeness. 
On the other hand, there is no obvious difference in testing accuracy with different \textcolor{blue}{sparsification ratios}, which indicates that the performance under the proposed DP-FCRN does not deteriorate much while reducing the communication burden.
%There is no significant difference in testing accuracy with different sparse coefficients. On the one hand, it can be shown that we can ensure that the performance does not deteriorate while reducing the communication burden.

\subsection{Trade-off between Privacy and Utility}
\begin{figure}[t]   
	\centering
	\includegraphics[width=1\linewidth]{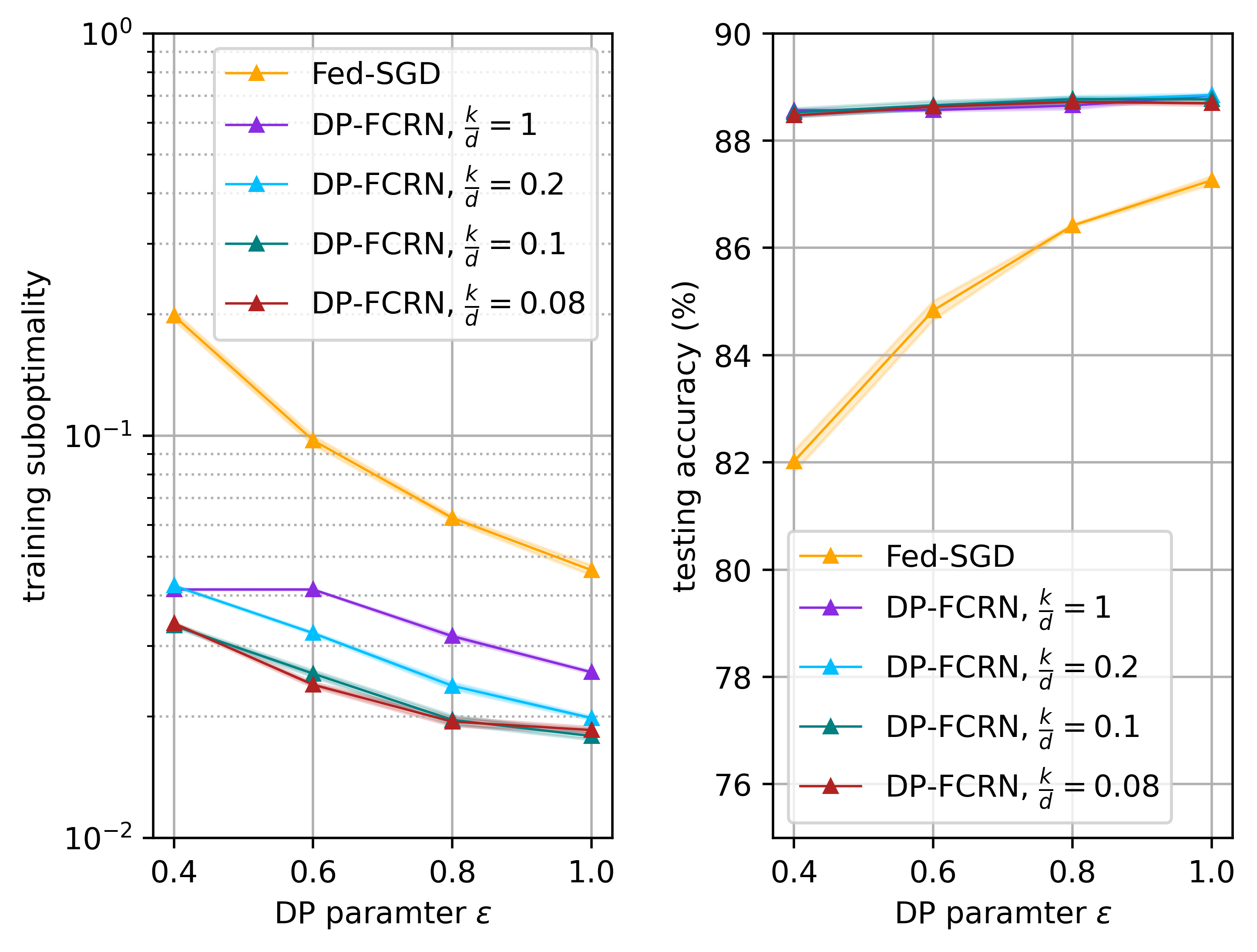}
	\caption{Performance comparison between Fed-SGD and DP-FCRN under different DP parameters.} 
	\label{fig: tradeoff}
\end{figure}

Fig.~\ref{fig: tradeoff} illustrates the trade-off between privacy and utility. It shows that when we increase the value of $\varepsilon$, i.e., relax the privacy requirement, the suboptimality will decrease across all the methods. 
Additionally, under a tighter DP requirement, i.e., smaller $\varepsilon$, the performance between DP-FCRN and Fed-SGD is more significant.

\section{Conclusion and Future Work} \label{sec: conclusions}
This paper explores communication efficiency and differential privacy within federated second-order methods. 
We demonstrate that the inherent sparsification characteristic can bolster privacy protection.
Moreover, employing second-order methods in a privacy setting can achieve the worst-case convergence guarantees and a faster convergence rate. 
Experiment results illustrate that our algorithm substantially outperforms first-order Fed-SGD in terms of utility loss.

There are several promising directions for future research. Firstly, investigating methods to reduce the computational complexity of federated second-order learning approaches is valuable. 
\textcolor{blue}{Additionally, exploring communication-efficient and privacy-preserving variants of advanced federated second-order algorithms, such as GIANT and SHED, presents promising research directions.}
%Additionally, integrating more general compression schemes and studying the privacy preservation and performance of non-convex cost functions are intriguing topics. 

%\bibliographystyle{unsrt}
%\bibliographystyle{automatica} 
\bibliographystyle{agsm}% Include this if you use bibtex 
\bibliography{HW}           % and a bib file to produce the 
% bibliography (preferred). The
% correct style is generated by
% Elsevier at the time of printing.

\appendix
\section*{Appendix}
\section{Supporting Lemmas}

We begin by introducing some properties of random-$k$ sparsification.

\begin{lemma}
	The random-$k$ sparsification operator $\mathcal{S}(x)$ exhibits the following properties:
	\begin{equation*}
		\mathbb{E}[\mathcal{S}(x)] = x, \ \mathbb{E}\left[ \| \mathcal{S}(x) - x \|^{2} \right] \leq \left(\frac{d}{k} - 1 \right)\|x \|^{2}.
	\end{equation*}
\end{lemma}

The following lemma provide some useful properties of $\phi_{i}(v;w)$~(\cite{ganesh2023faster}).
\begin{lemma} \label{lem: phi_pro}
	For any $i \in \{1, 2, \dots, n \}$, $\phi_{i}$ defined in~\eqref{eq: phi_1} has the following properties:
	\begin{itemize}
		\item[1)] For any $M \geq 0$ and $w, v \in \mathcal{X}, v \neq w$, there is
		\begin{align*}
			&	\nabla_{v}^{2} \phi_{i}(v;w) \\
			=& \nabla^{2} f_{i}(w) + \frac{M}{2}\|v-w\| I_{d} + \frac{M}{2\| v-w \|}(v-w)(v-w)^{T}.
		\end{align*}
		Therefore, $\nabla_{v}^{2} \phi_{i}(v;w) \succeq \lambda_{\min}(\nabla^{2}f_{i}(w))I_{d} + M\|v-w\|I_{d}$.
		\item[2)] For any $M \geq L_{2}$, and $v, w \in \mathcal{X}$,
		\begin{equation*}
			f_{i}(v) \leq \phi_{i}(v;w).
		\end{equation*}
		\item[3)] For any $M \geq 0$ and $v, w \in \mathcal{X}$,
		\begin{equation*}
			\phi_{i}(v;w) \leq f_{i}(v) + \frac{M + L_{2}}{6}\|v-w \|^{3}.
		\end{equation*}
	\end{itemize} 
\end{lemma}
It can be verified that $\phi(v;w) = \frac{1}{n}\sum_{i=1}^{n} \phi_{i}(v;w)$. Therefore, we can obtain similar properties between $\phi$ and $f$ as in Lemma~\ref{lem: phi_pro}.

\begin{lemma} \label{lem: induction}
For a sequence $\{ q_{t} \}_{t \geq 0}$ where $q_{t} \geq 1$ for all $t \geq 0$, if
\begin{equation*}
    {q}_{t+1} \leq {q}_{t} - \frac{1}{3} {q}_{t}^{\frac{3}{4}},
\end{equation*}
then
\begin{equation} \label{eq: ind_q}
	{q}_{t} \leq \left[ {q}_{0}^{\frac{1}{4}} - \frac{t}{12} \right]^{4}, \ \forall t \geq 0.
\end{equation}
\end{lemma}
\begin{pf}
    We prove Lemma~\ref{lem: induction} by induction. 
    For $t=0$, inequality~\eqref{eq: ind_q} is trivially true.
    Suppose~\eqref{eq: ind_q} holds for $t=k$, i.e., 
    \begin{equation*}
        {q}_{k} \leq \left[ {q}_{0}^{\frac{1}{4}} - \frac{k}{12} \right]^{4}.
    \end{equation*}
    Since the function $x - \frac{1}{3} x^{\frac{3}{4}}$ is increasing w.r.t. $x$, we have 
    \begin{align*}
        q_{k+1} \leq & {q}_{k} - \frac{1}{3} {q}_{k}^{\frac{3}{4}}  \\
        \leq & \left[ {q}_{0}^{\frac{1}{4}} - \frac{k}{12} \right]^{4} - \frac{1}{3} \left[ {q}_{0}^{\frac{1}{4}} - \frac{k}{12} \right]^{3}.
    \end{align*}
    To prove~\eqref{eq: ind_q} holds true for $t=k+1$, we need to show
    \begin{equation} \label{eq: ind_3}
         \left[ {q}_{0}^{\frac{1}{4}} - \frac{k}{12} \right]^{4} - \frac{1}{3} \left[ {q}_{0}^{\frac{1}{4}} - \frac{k}{12} \right]^{3} \leq \left[ {q}_{0}^{\frac{1}{4}} - \frac{k+1}{12} \right]^{4}.
    \end{equation}
Using the equality $a^{4}-b^{4} = (a-b)(a^{3} + a^{2}b + ab^{2} + b^{3})$, inequality~\eqref{eq: ind_3} is equivalent to
\begin{align*}
    \frac{1}{3} \left[ {q}_{0}^{\frac{1}{4}} - \frac{k}{12} \right]^{3} \geq &   \left[ {q}_{0}^{\frac{1}{4}} - \frac{k}{12} \right]^{4} - \left[ {q}_{0}^{\frac{1}{4}} - \frac{k+1}{12} \right]^{4} \\
    =& \frac{1}{12}\Bigg[  \left[ {q}_{0}^{\frac{1}{4}} - \frac{k+1}{12} \right]^{3} \\
    &+ \left[ {q}_{0}^{\frac{1}{4}} - \frac{k+1}{12} \right]^{2}\left[ {q}_{0}^{\frac{1}{4}} - \frac{k}{12} \right] \\
    &+ \left[ {q}_{0}^{\frac{1}{4}} - \frac{k+1}{12} \right]\left[ {q}_{0}^{\frac{1}{4}} - \frac{k}{12} \right]^{2} \\
    &+ \left[ {q}_{0}^{\frac{1}{4}} - \frac{k}{12} \right]^{3} \Bigg],
\end{align*}
which holds true since ${q}_{0}^{\frac{1}{4}} - \frac{k+1}{12} \leq {q}_{0}^{\frac{1}{4}} - \frac{k}{12}$. 
Thus, \eqref{eq: ind_3} is established, completing the induction proof of Lemma~\ref{lem: induction}.
\end{pf}

\begin{lemma} \label{lem: seq_1}
(\cite{ganesh2023faster})
Let $b_{0} > 0$ and define the sequence $a_{t+1} \leq b_{0} + \frac{1}{2}a_{t}^{\frac{3}{2}}$ where $a_{0} \leq \frac{16}{9}$. Then, after $T = \Theta(\log \log (\frac{1}{b}))$, we have $a_{T} = O(b_{0})$.
\end{lemma}

\section{Proof of Theorem~\ref{thm: dp}}
%\labelsubseccounter{app: thm_dp}
\label{app: thm_dp}

We first present some relevant properties of DP for privacy analysis.
\begin{lemma} \label{lem: subsample}
(Privacy for Subsampling~(\cite{steinke2022composition}))
	Suppose $\mathcal{G}$ is an $(\varepsilon, \delta)$-DP mechanism. 
 Consider $\text{\ttfamily Sample}_{r_{1}, r_{2}}: \mathcal{D}^{r_{1}} \to \mathcal{D}^{r_{2}}$ as the subsampling manipulation.
 Given a dataset belonging to $\mathcal{D}^{r_{1}}$ as an input, this subsampling manipulation selects a subset of $r_{2} \leq r_{1}$ elements from the input dataset uniformly at random. For the following mechanism
	\begin{equation*}
		\mathcal{G} \circ \text{\ttfamily Sample}_{r_{1},r_{2}}(D),
	\end{equation*} 
	where $D \in \mathcal{D}^{r_{1}}$. Then the mechanism $\mathcal{G} \circ \text{\ttfamily Sample}_{r_{1},r_{2}}$ is $(\varepsilon^{\prime}, \delta^{\prime})$-DP for $\varepsilon^{\prime} = \log(1+r_{2}(e^{\varepsilon} - 1)/r_{1})$ and $\delta^{\prime} = r_{2}\delta/r_{1}$.
\end{lemma}

\begin{lemma} \label{lem: compose}
	(Composition of DP~(\cite{steinke2022composition}))
	If each of $T$ randomized algorithms $\mathcal{A}_{1}, \dots, \mathcal{A}_{T}$ is $(\varepsilon_{i}, \delta_{i})$-DP with $\varepsilon_{i} \in (0, 0.9]$ and $\delta_{i} \in (0,1]$, then $\mathcal{A}$ with $\mathcal{A}(\cdot) = (\mathcal{A}_{1}(\cdot), \dots, \mathcal{A}_{T}(\cdot))$ is $(\tilde{\varepsilon}, \tilde{\delta})$-DP with
	\begin{equation*}
		\tilde{\varepsilon} = \sqrt{\sum_{t=1}^{T}2\varepsilon_{t}^{2} \log \left( e + \frac{\sqrt{\sum_{t=1}^{T} \varepsilon_{t}^{2}}}{\hat{\delta}} \right) } + \sum_{t=1}^{T} \varepsilon_{t}^{2}
	\end{equation*}
	and
	\begin{equation*}
		\tilde{\delta} = 1 - (1-\hat{\delta})\prod_{t=1}^{T}(1-\delta_{t})
	\end{equation*}
	for any $\hat{\delta} \in (0,1]$.
\end{lemma}

We first analyze DP at each local computation.
The Gaussian noise injected to each coordinate in $[\text{grad}_{i}^{t,s}]_{c_{i}^{t}}$ is generated from $\mathcal{N}(0, \sigma^{2})$. Then based on Lemma~\ref{lem: GM}, every local iteration in GMSolver preserves $(\varepsilon_{s}, \delta_{0})$-DP for each sampled data $\zeta_{i,t}$ with 
\begin{equation*}
	\varepsilon_{s} = \frac{2\sqrt{2k \log (1.25/\delta_{0})}{(\textcolor{blue}{L_{0}} + \textcolor{blue}{L_{1}}D)}}{\sigma \sqrt{d}}
\end{equation*} 
for any $\delta_{0} \in [0,1]$. 

Based on Lemma~\ref{lem: subsample}, each local iteration of GMSolver preserves $(\varepsilon_{s}^{\prime}, \delta_{0}/m)$-DP for client $i$'s local dataset $\zeta_{i}$, where 
\begin{equation*}
	\varepsilon_{s}^{\prime} = \log \left( 1+ \frac{e^{\varepsilon_{s}} - 1}{m} \right) \leq \frac{2 \varepsilon_{s}}{m}.
\end{equation*}
According to the conditions on $T$ and $\sigma$ shown in Theorem~\ref{thm: dp}, we have
\begin{equation*}
	\varepsilon_{s}^{\prime 2} \leq  \frac{32k\log(1.25/\delta_{0})(L_{0} + L_{1}D)^{2}}{\sigma^{2}m^{2}d} \leq \frac{\varepsilon^{2}}{5\tau T} \leq 0.8.
\end{equation*}
Therefore, we have $\varepsilon_{s}^{\prime} \leq 0.9$ and
\begin{equation} \label{eq: sum_ep}
	\sum_{s=1}^{\tau T} \varepsilon_{s}^{\prime 2} \leq \frac{1}{5} \sum_{s=1}^{\tau T} \frac{\varepsilon^{2}}{\tau T} \leq 1
\end{equation}
for the given $\varepsilon \in (0,1]$.

Then we analyze DP after $T$ iterations.
After performing $T$ communication rounds, client $i$ conducts $T\tau$ iterations of local computation.
Therefore, using Lemma~\ref{lem: compose}, we obtain DP-FCRN obtains $(\tilde{\varepsilon}, \tilde{\delta})$-DP with
\begin{equation*}
	\tilde{\varepsilon} = \sqrt{\sum_{s=1}^{\tau T}2\varepsilon_{s}^{\prime 2} \log \left( e + \frac{\sqrt{\sum_{s=1}^{\tau T} \varepsilon_{s}^{\prime 2}}}{\tilde{\delta}} \right) } + \sum_{s=1}^{ \tau T} \varepsilon_{s}^{\prime 2}
\end{equation*}
and $\tilde{\delta} = 1-(1-\delta^{\prime})(1-\delta_{0}/m)^{\tau T}$ for any $\delta^{\prime} \in (0,1]$. 
Furthermore, there is
\begin{align*}
	\tilde{\varepsilon} \leq & \sqrt{\sum_{s=1}^{\tau T}2\varepsilon_{s}^{ \prime 2} \log \left( e + \frac{\sqrt{\sum_{s=1}^{\tau T} \varepsilon_{s}^{\prime 2}}}{\tilde{\delta}} \right) } + \frac{1}{5}\varepsilon^{2} \\
	\leq & \sqrt{3\sum_{s=1}^{\tau T}\varepsilon_{s}^{\prime 2}} + \frac{1}{5} \varepsilon \\
	\leq & \sqrt{\frac{3}{5}\varepsilon^{2}} + \frac{1}{5}\varepsilon \\
	\leq & \varepsilon,
\end{align*}
where the second inequality holds from~\eqref{eq: sum_ep}. If we set $\delta^{\prime} = \sqrt{\sum_{s=1}^{\tau T} \varepsilon_{s}^{2}}$ and $\delta =\tilde{\delta}$, the we have DP-FCRN preserves $(\varepsilon, \delta)$-DP.

\section{Proof of Lemma~\ref{lem: solver}}
%\labelsubseccounter{app: lem_solver}
\label{app: lem_solver}
We can write a stochastic estimate of $\phi_{i}(v;w)$ as follows:
\begin{align*}
		&\hat{\phi}_{i}(v;w) \\
\triangleq & f(w) + \left< \hat{g}_{i}, v-w \right> + \frac{1}{2}\left< \hat{H}_{i}(v-w), v-w \right> \\
& + \frac{M}{6}\|v-w \|^{3}, 
\end{align*}
where $\hat{g}_{i}$ and $\hat{H}_{i,t}$ are stochastic estimates of $\nabla f_{i}(w)$ and $\nabla^{2}f_{i}(w)$.
According to Algorithm~\ref{algo: one}, we find that $\text{grad}_{s}$ is a stochastic gradient of $\nabla_{\theta_{s}} \phi(\theta_{s}, \theta_{0})$. Based on the non-expansive property of the projection operator, we have
\begin{align*}
	&\mathbb{E}\left[\| \theta_{s+1} - \theta_{*} \|^{2} | \mathcal{F}_{s} \right]\\
	 \leq & \| \theta_{s} - \theta_{*} \|^{2} + \eta_{s}^{2} \mathbb{E}\left[ \|\text{grad}_{s} + b_{s} \|^{2} | \mathcal{F}_{s} \right] \\
	& - 2\eta_{s} \left<\nabla_{\theta_{s}}\phi_{i}(\theta_{s};\theta_{0}), \theta_{s} - \theta_{0} \right> \\
	\leq & \| \theta_{s} - \theta_{*} \|^{2} + \eta_{s}^{2} \mathbb{E}\left[ \|\text{grad}_{s} + b_{s} \|^{2}| \mathcal{F}_{s} \right] \\ 
	& -2\eta_{s}\left[\phi_{i}(\theta_{s};\theta_{0}) - \phi_{i}(\theta^{*};\theta_{0}) + \frac{\mu}{2}\|\theta_{s};\theta_{0} \|^{2} \right ]  ,
\end{align*}
where the last inequality holds from the $\mu$-strong convexity of $\nabla \phi_{i}$. By arranging the inequality, we have
\begin{align} \label{eq: solver_ite}
	& \mathbb{E}[\phi_{i}(\theta_{s}; \theta_{0})] - \phi_{i}(\theta^{*};\theta_{0}) \nonumber \\
	\leq & \frac{\eta_{s}(L^{2} + \sigma^{2}d)}{2} + \left( \frac{1}{2\eta_{s}}- \frac{\mu}{2} \right) \mathbb{E}[\| \theta_{s} - \theta^{*} \|^{2}] \nonumber \\
	& - \frac{1}{2\eta_{s}} \mathbb{E}[\| \theta_{s+1} - \theta^{*} \|^{2}],
\end{align}
where $L = L_{0} + L_{1}D + \frac{M}{2}D^{2}$.
With $\eta_{s} = \frac{2}{\mu(s+2)}$ and multiplying the~\eqref{eq: solver_ite} by $s+1$, we obtain
\begin{align*}
	&(s+1) \left( \mathbb{E}[\phi_{i}(\theta_{s}; \theta_{0})] - \phi_{i}(\theta^{*};\theta_{0}) \right) \nonumber \\
	\leq & \frac{(s+1)(L^{2} + \sigma^{2}d)}{\mu (s+2)} - \frac{\mu(s+2)(s+1)}{4}\mathbb{E}[\| \theta_{s+1} - \theta^{*} \|^{2}] \\
	& + \left( \frac{\mu(s+2)(s+1)}{4}  - \frac{\mu(s+1)}{2}\right)\mathbb{E}[\| \theta_{s} - \theta^{*} \|^{2}] \\
	\leq & \frac{L^{2} + \sigma^{2}d}{\mu} + \frac{\mu}{4}\bigg[s(s+1)\mathbb{E}\left [\| \theta_{s} - \theta^{*} \|^{2} \right] \\
	& \quad \quad \quad - (s+1)(s+2)\mathbb{E} \left[\| \theta_{s+1} - \theta^{*} \|^{2} \right]  \bigg].
\end{align*}
By summing from $s=0$ to $s=\tau$ of these $s$-weighted inequalities, we have
\begin{align*}
	&\sum_{s=0}^{\tau-1} (s+1) \left( \mathbb{E}[\phi_{i}(\theta_{s}; \theta_{0})] - \phi_{i}(\theta^{*};\theta_{0}) \right) \\
	\leq & \frac{\tau(L^{2} + \sigma^{2}d)}{\mu} - \frac{\mu}{4}\tau(\tau+1)\mathbb{E} \left[\| \theta_{\tau} - \theta^{*} \|^{2} \right].
\end{align*}
Thus, 
\begin{align*}
	& \mathbb{E} \left[ \phi_{i}\left(\frac{2}{\tau(\tau+1)}\sum_{s=0}^{\tau-1} (s+1)\theta_{s};\theta_{0} \right) \right] - \phi_{i}(\theta^{*};\theta_{0}) \\
 \leq & \frac{2(L^{2} + \sigma^{2}d)}{\mu(\tau+1)}.
\end{align*}

%In each step, we consider $\text{grad}_{s} + b_{s}$ for the noisy gradient. 

%From Lemma 1 in~\cite{jin2019short}, we know that $\|b_{s}\|$ is a SubGaussian random variable with variance proxy of $c \sigma \sqrt{d}$, where $c$ is a universal constant.
%Based on Theorem C.3 in~\cite{harvey2019simple}, we can obtain that for every $\beta \in (0,1]$, the suboptimality gap with probability at least $1-\beta$ is given by
%\begin{equation*}
%	O\left( \frac{(L + \sigma\sqrt{d})^{2}}{\mu^{2}} \cdot \frac{\log(1/ \beta)}{\tau} \right) = O\left( \frac{L^{2} + \sigma^{2}d}{\mu^{2}} \cdot \frac{\log(1/ \beta)}{\tau} \right),
%\end{equation*}
%where we use $(a+b)^{2} \leq 2a^{2} + 2b^{2}$ for every $a$ and $b$, and $L = L_{0} + L_{1}D + \frac{M}{2}D^{2}$.
Therefore, after the local computation of Algorithm~\ref{algo: two}, the suboptimality gap is given by
\begin{equation*}
 O\left( \frac{L^{2} + \sigma^{2}d}{\mu \tau}  \right).
\end{equation*}
Putting the value of $\sigma$ in~\eqref{eq: noise} obtains:
\begin{equation*}
	O\left(  \left( \frac{L^{2}}{\mu \tau} + \frac{k T\log(1/\delta_{0}){(\textcolor{blue}{L_{0}} + \textcolor{blue}{L_{1}}D)}^{2}}{\varepsilon^{2}m^{2}\mu } \right) \right).
\end{equation*}
Then, by setting the number of local iterations to $\tau = \frac{L^{2}\varepsilon^{2}m^{2}}{kT\log(1/\delta_{0}) (\textcolor{blue}{L_{0}} + \textcolor{blue}{L_{1}}D)^{2}  }$, we obtain that the subotimality is given by~\eqref{eq: solver_gap}.

\section{Proof of Theorem~\ref{thm: convergence}}
%\labelsubseccounter{app: thm_con}
\label{app: thm_con}
Using 2) in Lemma~\ref{lem: phi_pro}, we can write
\begin{align} \label{eq: diff_1}
&\mathbb{E}[ f(x_{t+1})] - f(x^{*}) \nonumber \\
\leq & \mathbb{E}[ \phi(x_{t+1};x_{t})] - f(x^{*}) \nonumber \\
=& \mathbb{E}\left[ \phi(x_{t+1};x_{t}) - \frac{1}{n}\sum_{i=1}^{n} \phi_{i}(x_{i,t+1};x_{t}) + \frac{1}{n}\sum_{i=1}^{n} \phi_{i}(x_{i,t+1};x_{t}) \right] \nonumber \\
& - \frac{1}{n}\sum_{i=1}^{n} \min_{x^{(i)} \in \mathcal{X}}\phi_{i}(x^{(i)};x_{t}) + \frac{1}{n}\sum_{i=1}^{n} \min_{x^{(i)} \in \mathcal{X}}\phi_{i}(x^{(i)};x_{t}) \nonumber \\
& -f(x^{*}) \nonumber \\
\leq & \frac{1}{n}\sum_{i=1}^{n} \left[ \mathbb{E}[ \phi_{i}(x_{i,t+1};x_{t})]- \min_{x^{(i)} \in \mathcal{X}}\phi_{i}(x^{(i)};x_{t})  \right] \nonumber \\
& +\mathbb{E} \left[ \phi(x_{t+1};x_{t}) - \frac{1}{n}\sum_{i=1}^{n} \phi_{i}(x_{i,t+1};x_{t}) \right] \nonumber \\
& + \left[\min_{x \in \mathcal{X}} \phi(x;x_{t}) - f(x^{*}) \right],
\end{align}
where the last inequality uses the fact that
\begin{equation*}
    \frac{1}{n}\sum_{i=1}^{n} \min_{x^{(i)} \in \mathcal{X}}\phi_{i}(x;x_{t}) \leq \min_{x \in \mathcal{X}} \phi(x;x_{t}).
\end{equation*}
%$\frac{1}{n}\sum_{i=1}^{n} \min_{x^{(i)} \in \mathcal{X}}\phi_{i}(x;x_{t}) \leq \min_{x \in \mathcal{X}} \phi(x;x_{t})$.
%\begin{equation*} 
%	\textcolor{blue}{\frac{1}{n}\sum_{i=1}^{n} \min_{x^{(i)} \in \mathcal{X}}\phi_{i}(x;x_{t}) \leq \min_{x \in \mathcal{X}} \phi(x;x_{t})}
%\end{equation*}
Since $\mathcal{X}$ is a closed and convex set and $\phi(x;x_{t})$ is a strongly convex function w.r.t. $x$, we conclude that there exists a unique $x_{t+1}^{*} = \arg\min_{x \in \mathcal{X}} \phi(x;x_{t})$.

At each $t$, we obtain an approximate minimizer of $\phi_{i}(x;x_{t})$ based on the GMSolver:
\begin{equation*}
\begin{aligned}
	 &\frac{1}{n}\sum_{i=1}^{n} \left[ \mathbb{E}[ \phi_{i}(x_{i,t+1};x_{t})]- \min_{x^{(i)} \in \mathcal{X}}\phi_{i}(x^{(i)};x_{t})  \right] \\
	\leq & O\left( \frac{k \log(1 / \delta_{0}) {(\textcolor{blue}{L_{0}} + \textcolor{blue}{L_{1}}D)}^{2}T}
		{\varepsilon^{2}m^{2}\mu } \right) \triangleq \Gamma_{1}.
\end{aligned}
\end{equation*}
Lemma~\ref{lem: solver} provides the performance guarantee of the GMSolver and shows that at each step of Algorithm~\ref{algo: one}, the optimization error in minimizing $\phi_{i}(x^{(i)};x_{t})$ is less than $\Gamma_{1}$. 

%For the first term of~\eqref{eq: diff_1}, we have the following result according to Lemma~\ref{lem: solver}.
%\begin{align*}
%	O\left( \frac{k \log(1 / \delta_{0}) (L_{0} + L_{1}D)^{2}T}
%		{\varepsilon^{2}m^{2}d\mu } \cdot \log(1/\beta) \triangleq \Gamma_{1}
% \right) 
%\end{align*} 

For the second term of~\eqref{eq: diff_1}, we obtain
\begin{align} \label{eq: Gamma_2}
	&\mathbb{E}\left[ \phi(x_{t+1};x_{t}) - \frac{1}{n}\sum_{i=1}^{n} \phi_{i}(x_{i,t+1};x_{t}) \right] \nonumber \\
	=& \frac{1}{n}\sum_{i=1}^{n} \mathbb{E} \left[ \phi_{i}(x_{t+1};x_{t}) - \phi_{i}(x_{i,t+1};x_{t}) \right] \nonumber \\
	\leq & \frac{1}{n}\sum_{i=1}^{n} \mathbb{E}[ \nabla_{x_{t+1}}\phi_{i}(x_{t+1};x_{t})(x_{t+1} - x_{i,t+1}) ] \nonumber \\
	=& \frac{1}{n}\sum_{i=1}^{n} \nabla_{x_{t+1}}\phi_{i}(x_{t+1};x_{t}) \mathbb{E}[ (x_{t+1} - x_{t}) - (x_{i,t+1} - x_{t}) ] \nonumber \\
	=& \frac{1}{n}\sum_{i=1}^{n} \nabla_{x_{t+1}}\phi_{i}(x_{t+1};x_{t}) \mathbb{E}\left[ \frac{1}{n} \sum_{i=1}^{n}\mathcal{S}_{i}^{t}(y_{i,t}) - \frac{y_{i,t}}{\alpha} \right] \nonumber \\
	=& \frac{1}{n}\sum_{i=1}^{n} \nabla_{x_{t+1}}\phi_{i}(x_{t+1};x_{t}) \left( \frac{1}{n} \sum_{i=1}^{n}y_{i,t} - \frac{y_{i,t}}{\alpha} \right) \nonumber \\
	\leq & \frac{1}{n} \sum_{i=1}^{n} L \left( \frac{1}{n} \sum_{i=1}^{n}\|y_{i,t}\|  
 + \| y_{i,t} \| \right) \nonumber \\  %\triangleq \alpha \Gamma_{2}, 
 \leq & 2 \alpha LD,
\end{align}
where the first inequality follows from the Lipschitz continuous of $\phi_{i}$, the third equality holds from steps 8 and 9 in Algorithm~\ref{algo: one}, and the second inequality holds from $\alpha > 1$.
Putting 
\begin{equation*}
	\alpha = O \left( \frac{\Gamma_{1}}{LD} \right)
\end{equation*}
into~\eqref{eq: Gamma_2}, we have
\begin{align*}
	& \phi(x_{t+1};x_{t}) - \frac{1}{n}\sum_{i=1}^{n} \phi_{i}(x_{i,t+1};x_{t}) \leq O(\Gamma_{1}).
\end{align*}
Then, we provide an upper bound on the last term of~\eqref{eq: diff_1}. We obtain the following relationship by 3) in Lemma~\ref{lem: phi_pro}.
\begin{align*}
	& \min_{x \in \mathcal{X}} \phi(x;x_{t}) - f(x^{*}) \\
	\leq & \min_{x \in \mathcal{X}} \left[ f(x) + \frac{M + L_{2}}{6}\|x - x_{t}\|^{3} - f(x^{*}) \right].
\end{align*}
Since $\mathcal{X}$ is a convex set and $x_{t}, x^{*} \in \mathcal{X}$, for all $\eta \in [0,1]$, $(1-\eta)x_{t} + \eta x^{*} \in \mathcal{X}$. Therefore,
\begin{align*}
	& \min_{x \in \mathcal{X}} \left[ f(x_{t}) + \frac{M + L_{2}}{6}\|x - x_{t}\|^{3} - f(x^{*}) \right] \\
\leq & \min_{\eta_{t} \in [0,1]} \bigg[ f\big((1-\eta_{t})x_{t} + \eta_{t}x^{*} \big) + \eta_{t}^{3}\frac{M+L_{2}}{6}\| x_{t} - x^{*} \|^{3} \\
& \quad \quad \quad \quad - f(x^{*}) \bigg].
\end{align*}
By the convexity of $f$, we have $f((1-\eta_{t})x_{t} + \eta_{t}x^{*}) \leq f(x_{t})  - \eta_{t}\left(f(x_{t}) - f(x^{*}) \right)$.
Also, strong convexity implies that $\|x_{t+1} - x^{*} \|^{3} \leq \left[ \frac{2}{\mu }( f(x_{t}) - f(x^{*} ) \right]^{\frac{3}{2}}$. Thus,
\begin{align} \label{eq: term_3}
	& \min_{x \in \mathcal{X}} \phi(x;x_{t}) - f(x^{*}) \nonumber \\
\leq & \min_{\eta_{t} \in [0,1]} \Bigg\{ f(x_{t}) - f(x^{*}) - \eta_{t}( f(x_{t}) - f(x^{*}) ) \nonumber \\
& \quad \quad \quad \quad \quad + \eta_{t}^{3}\frac{M+L_{2}}{6} \left[ \frac{2}{\mu }( f(x_{t}) - f(x^{*} ) \right]^{\frac{3}{2}} \Bigg \}.
\end{align}
Let $\lambda = \Big( \frac{3}{M+L_{2}} \Big)^{2} \Big( \frac{\mu}{2} \Big)^{3}$ and $u_{t} = \lambda^{-1}\big( f(x_{t}) - f(x^{*}) \big)$. %Then, 
%under the event $E$ and 
Based on~\eqref{eq: term_3}, we can rephrase~\eqref{eq: diff_1} as
\begin{equation} \label{eq: q_iteration}
	u_{t+1} \leq \lambda^{-1} \Gamma_{1} + \min_{\eta_{t} \in [0,1]} \left( u_{t} - \eta_{t}u_{t} + \frac{1}{2} \eta_{t}^{3}u_{t}^{\frac{3}{2}}  \right).
\end{equation}
Denote $\eta_{t}^{*} = \arg\min_{\eta_{t} \in [0,1]} \left( u_{t} - \eta_{t}u_{t} + \frac{1}{2} \eta_{t}^{3}u_{t}^{\frac{3}{2}}  \right)$, we have that $\eta_{t} = \min \left \{\sqrt{\frac{2}{3\sqrt{u_{t}}}}, 1 \right \}$.

We have two convergence cases according to different choices of $\eta^{*}$.

\textbf{Phase \uppercase\expandafter{\romannumeral1}}: If $u_{t} \geq \frac{4}{9}$, then $\eta^{*} = \sqrt{\frac{2}{3\sqrt{u_{t}}}}$. The iteration~\eqref{eq: q_iteration} will become
\begin{equation*}
	u_{t+1} \leq \lambda^{-1}\Gamma_{1} + u_{t} - \left(\frac{2}{3} \right)^{\frac{3}{2}}u_{t}^{\frac{3}{4}}.
\end{equation*}
%%%%%%%%%%%%%%%%%%%
\textbf{Phase \uppercase\expandafter{\romannumeral2}}: If $u_{t} < \frac{4}{9}$, then $\eta^{*} = 1$. The iteration~\eqref{eq: q_iteration} will be given by
\begin{equation*}
	u_{t+1} \leq \lambda^{-1}\Gamma_{1} + \frac{1}{2}u_{t}^{\frac{3}{2}}.
\end{equation*}
Assume that $u_{0} \geq \frac{4}{9}$.
In the following analysis, we will show that, $\{u_{t} \}_{t \in [T]}$ is a decreasing sequence. Therefore, we can conclude that there exists a time step $T_{1} > 0$, %\textcolor{blue}{independent of xx}, 
such that $u_{t} < \frac{4}{9}$ for $t \geq T_{1}$. Subsequently, for $t \geq T_{1}$, there will be $\eta_{t}^{*} = 1$.

%%%%%%%%%%%%%%%%% 
For the convergence of Phase \uppercase\expandafter{\romannumeral1}, inspired by~\cite{nesterov2006cubic}, we let $\tilde{u}_{t+1} = \frac{9}{4}u_{t}$, and assume $u_{t} \geq \frac{3\Gamma_{1}}{\lambda}$.
%$\Gamma_{1} \leq \frac{4\lambda}{27}$. 
Then, there is $\tilde{u}_{t+1} \geq 1$ and the evolution of Phase \uppercase\expandafter{\romannumeral1} becomes:
\begin{align} \label{eq: induction}
	\tilde{u}_{t+1} \leq & \frac{9\Gamma_{1}}{4\lambda} + \tilde{u}_{t} - \frac{2}{3} \tilde{u}_{t}^{\frac{3}{4}} \nonumber \\
	\leq & \tilde{u}_{t} - \frac{1}{3}\tilde{u}_{t}^{\frac{3}{4}},
\end{align}
where the last inequality holds from $\frac{9 \Gamma_{1}}{4 \lambda} \leq \frac{\tilde{u}_{t}^{\frac{3}{4}}}{3}$.
% $\tilde{u}_{t} \geq 1$ and $\frac{9 \Gamma_{1}}{4 \lambda} \leq \frac{1}{3} \leq \frac{\tilde{u}_{t}^{\frac{3}{4}}}{3}$.
%.`When $\tilde{u}_{t} \geq 1$, there is $\tilde{u}_{t+1} \leq \tilde{u}_{t}$.
According to Lemma~\ref{lem: induction}, we have
\begin{equation} \label{eq: ind_pro}
	\tilde{u}_{t} \leq \left[ \tilde{u}_{0}^{\frac{1}{4}} - \frac{t}{12} \right]^{4}, 
\end{equation}
which indicates
\begin{equation*}
	\frac{9{u}_{t}}{4} \leq \left[ \left( \frac{9{u}_{0}}{4} \right)^{\frac{1}{4}} - \frac{t}{12} \right]^{4}.
\end{equation*}
To make $u_{T_{1}^{*}} < \frac{4}{9}$, there is
\begin{equation*}
	\frac{9{u}_{T_{1}^{*}}}{4} \leq \left[ \left( \frac{9{u}_{0}}{4} \right)^{\frac{1}{4}} - \frac{T_{1}^{*}}{12} \right]^{4} \leq \frac{4}{9},
\end{equation*}
which implies that 
\begin{equation} \label{eq: phase_1}
	T_{1}^{*} = O \left( \frac{\sqrt{M + L_{2}}( f(x_{0}) - f(x^{*}) )^{\frac{1}{4}}}{\mu^{\frac{3}{4}}} \right).
\end{equation}
Therefore, after $T_{1}^{*}$ iterations, we enter Phase \uppercase\expandafter{\romannumeral2}.

%%%%%%%%%%%%%%%%%%%%%%
For the convergence analysis of phase \uppercase\expandafter{\romannumeral2},
the evolution is given by
\begin{equation*}
	u_{t+1} \leq \lambda^{-1} \Gamma_{1} + \frac{1}{2}u_{t}^{\frac{3}{2}}.
\end{equation*}
%Using Lemma~\ref{lem: seq_1}, we obtain that after 
We define another sequence $\{w_{t}\}_{t \geq 0}$, with $w_{0} = u_{0}, \ w_{t+1} = \frac{3}{4}(w_{t})^{\frac{3}{2}}$.
%as follows:
% \begin{equation*}
% 	r_{0} = u_{0}, \ r_{t+1} = \frac{3}{4}(r_{t})^{\frac{3}{2}}.
% \end{equation*}
By induction, we derive for every $t \geq 0$ where $\lambda^{-1}\Gamma_{1} \leq \frac{1}{4}w_{t}^{\frac{3}{2}}$, there is $w_{t+1} \geq u_{t+1}$. Then, we can write
\begin{equation*}
	w_{t+1} = \frac{3}{4}w_{t}^{\frac{3}{2}}, \ \frac{9}{16}w_{t+1} = \left( \frac{9}{16}w_{t} \right)^{\frac{3}{2}}.
\end{equation*}
Therefore, we obtain that $\log(\frac{9}{16}w_{t}) = (\frac{3}{2})^{t}\log(\frac{9}{16}w_{0})$.
We want to find $T$ such that $\lambda^{-1}\Gamma_{1} \leq \frac{1}{4}w_{T}^{\frac{3}{2}} \leq 2\lambda^{-1}\Gamma_{1} $, i.e., $\frac{2}{3}\log(\frac{27}{16}\lambda^{-1} \Gamma_{1}) \leq \log( \frac{9}{16} w_{T}) \leq \frac{2}{3} \log( \frac{27}{8} \lambda^{-1}\Gamma_{1})$. Hence, we obtain $T = \Theta\left(\log \Big(\log\big(\frac{\lambda}{\Gamma_{1} }\big) \Big) \right)$. 
As a result, there is $w_{T+1} = O(\lambda^{-1}\Gamma_{1} )$ and $u_{T+1} \leq w_{T+1} = O(\lambda^{-1}\Gamma_{1} )$.
Therefore, in Phase \uppercase\expandafter{\romannumeral2}, with the number of iterations
\begin{equation} \label{eq: phase_2}
	T_{2}^{*} = \tilde{\Theta}\left( \log \log\left( \frac{\varepsilon m }{\sqrt{k\log(1/\delta_{0})}} \right) \right),
\end{equation}
we obtain the best optimization error.

In summary, the optimization error is given by
\begin{align*}
	&f(x_{T}) - f(x^{*}) \\
	=& O\left( \frac{k \log(1 / \delta_{0}) {(\textcolor{blue}{L_{0}} + \textcolor{blue}{L_{1}}D)}^{2}}
		{\varepsilon^{2}m^{2}\mu } \cdot T \right),
\end{align*}
where $T = T_{1}^{*} + T_{2}^{*}$.
%and $T_{1}^{*}$ and $T_{2}^{*}$ are given by~\eqref{eq: phase_1} and~\eqref{eq: phase_2}, respectively.

\end{document}